\documentclass{article}

\usepackage{arxiv}

\usepackage[utf8]{inputenc} 
\usepackage[T1]{fontenc}    
\usepackage{hyperref}       
\usepackage{url}            
\usepackage{booktabs}       
\usepackage{amsfonts}       
\usepackage{nicefrac}       
\usepackage{microtype}      
\usepackage{lipsum}		
\usepackage{graphicx}
\usepackage[numbers]{natbib}
\usepackage{doi}
\usepackage{amsmath}
\usepackage{rotating}
\usepackage{cleveref}
\usepackage{todonotes}
\usepackage{subcaption}
\usepackage{lscape}
\usepackage{xcolor}
\usepackage[table]{xcolor}
\usepackage{ulem}

\usepackage{makecell}
\usepackage{afterpage}
\usepackage{siunitx}
\usepackage{float}
\usepackage{easyReview}
\usepackage{algorithm}
\usepackage{algpseudocode}
\usepackage{enumitem}
\numberwithin{equation}{section}
\algnewcommand\algorithmicinput{\textbf{Input:}}
\algnewcommand\algorithmicoutput{\textbf{Output:}}
\algnewcommand\Input{\item[\algorithmicinput]}
\algnewcommand\Output{\item[\algorithmicoutput]}

\setreviewson

\def\*#1{\mathbf{#1}}

\title{Adaptive Domain Decomposition for Finite Basis Physics-Informed Neural Networks}

\date{} 					

\author{
 \href{https://orcid.org/0009-0002-1337-0089}{\includegraphics[scale=0.06]{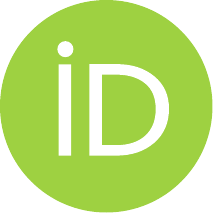}\hspace{1mm}Sarah Helfert}\\
             University of North Carolina, Charlotte\\
	Charlotte, NC USA \\
 	Pacific Northwest National Laboratory\\
	Richland, WA 99354  USA \\
     \And
 \href{https://orcid.org/0009-0008-6356-4002}{\includegraphics[scale=0.06]{orcid.pdf}\hspace{1mm}Wanxin Chen}\\
 	 Delft University of Technology\\
	 Delft Institute of Applied Mathematics \\
	 2628 CD Delft, The Netherlands \\
 \And
        \href{https://orcid.org/0000-0002-6411-6198}{\includegraphics[scale=0.06]{orcid.pdf}\hspace{1mm}Amanda A. Howard}\\
	Pacific Northwest National Laboratory\\
	Richland, WA 99354 USA \\
	\texttt{amanda.howard@pnnl.gov} \\
    \And
        \href{https://orcid.org/0000-0002-9928-5637}{\includegraphics[scale=0.06]{orcid.pdf}\hspace{1mm}Panos Stinis} \\
	Pacific Northwest National Laboratory\\
	Richland, WA 99354 USA \\
 	University of Washington, Applied Mathematics\\
	Seattle, WA USA\\
 	Brown University, Applied Mathematics\\
	  Providence, RI, 02912 USA \\
        \texttt{panagiotis.stinis@pnnl.gov}
            \And
         \href{https://orcid.org/0000-0003-1578-8104}{\includegraphics[scale=0.06]{orcid.pdf}\hspace{1mm}Alexander Heinlein}\\
	 Delft University of Technology\\
	 Delft Institute of Applied Mathematics \\
	 2628 CD Delft, The Netherlands \\
	 \texttt{a.heinlein@tudelft.nl}
}

\renewcommand{\shorttitle}{Adaptive Domain Decomposition for Finite Basis Physics-Informed Neural Networks}

\begin{document}
\maketitle

\begin{abstract}
    In this work, we present adaptive domain decomposition finite basis physics-informed neural networks (ADD-FBPINNs). The proposed method integrates Partition of Unity Networks (POUNets), a data-driven approach for learning domain partitions, with finite Basis physics-informed neural networks (FBPINNs) to perform more efficient domain decomposition and achieve more accurate approximations. FBPINNs improve upon the accuracy of standard physics-informed neural networks (PINNs) through user-defined subdivisions of the computational domain. By adapting the domain decomposition based on the solution's features using a separable POUNet, the framework can automatically learn partitions that are stable and can be geometrically interpretable. We test this approach on the two-dimensional Helmholtz equation, two-dimensional Poisson equation with multiscale features, and viscous Burgers equation and observe improved accuracy compared with FBPINNs with fixed domain decomposition.
\end{abstract}


\section{Introduction}

Scientific machine learning (SciML) is a rapidly developing field of research that combines scientific understanding with machine learning to produce accurate, robust, and efficient numerical methods for scientific computing. Modeling physical systems relies heavily on solving partial differential equations (PDEs). Classical numerical approaches, including finite difference, finite element, and spectral methods, have been used reliably to solve forward and inverse PDE problems \cite{ames2014book}. However, these methods still face inherent challenges, including difficulty in incorporating noisy or sparse data and the curse of dimensionality. One area of SciML research aims to bridge this gap by incorporating physical constraints into learning-based methods \cite{karniadakis2021physics}.

One popular approach is to use physics-informed neural networks (PINNs) \cite{raissi2017pipart1, raissi2017pipart2, raissi2019physics}, which directly enforce the governing equations of a PDE by incorporating the residual of the PDE into the loss function of a neural network.  As a result, PINNs offer mesh-free learning that can accurately approximate the solution to PDEs with little to no solution or measurement data required. PINNs have been shown to be effective in a variety of applications, including fluid dynamics and turbulence \cite{jin2021, cai2021physicsinformedneuralnetworkspinns, hanrahan2023, patel2024turbulencemodelaugmentedphysics, toscano2025pinns}. However, PINNs also suffer from several limitations, including spectral bias \cite{khodakarami2026spectral}, the need to tune hyperparameters, and poor scalability to large or complex domains and multiscale problems \cite{wang2020fail, wang2021spec, rahaman2019spectral}.

In particular, scaling PINNs to geometrically complex domains and high-frequency solutions remains a key barrier to their adoption in practical applications for several reasons. First, PINNs balance a multi-objective loss function that simultaneously enforces the PDE residual, boundary, and initial conditions. As the domain grows, this optimization becomes increasingly difficult, because more collocation points are required to adequately sample the domain. Adaptive resampling of collocation points and adaptive reweighting of loss terms have been proposed as methods to address this issue \cite{chen2025self, anagnostopoulos2024residual, mcclenny2023, wu2023rad, visser2026pacmann}. Second, the spectral bias present in neural networks further complicates training PINNs in complex domains. Spectral bias is the phenomenon that results in PINNs learning low-frequency features much faster than high-frequency features \cite{rahaman2019spectral, wang2021spec}, and larger domains mapped to a fixed input range effectively pack more oscillations of the solution per unit of network input, which may push low-frequency problems into the high-frequency regime. Spectral bias also makes learning solutions that exhibit high-frequency and multiscale features more difficult \cite{wang2021spec, khodakarami2026spectral}.

Domain decomposition (DD) methods have been combined with machine learning, including PINNs, in many studies; see \cite{heinlein2021dd,klawonn2024dd} for overviews. Early approaches, such as DeepDDM~\cite{li2020deepddm} and D3M~\cite{li2019d3m}, use neural networks to solve local subdomain problems within an outer Schwarz iteration, using residual-based or variational formulations. Related developments include nonoverlapping domain decomposition methods~\cite{jeon2026nnti, sun2024ddlearning} and two-level DeepDDM with a coarse network to improve scalability~\cite{dolean2024twoleveldeepddm}.

In this work, we focus on domain decomposition-based PINNs whose local networks are coupled through a global loss function and trained using gradient-based optimization, without an outer fixed-point iteration. This ``divide-and-conquer'' approach can simplify the optimization problem \cite{heinlein2021dd,klawonn2024dd}. Separate networks are used to learn pieces of the solution on partitions of the computational domain. One notable method is conservative PINNs (cPINNs) \cite{jagtap2020cpinns}, which explicitly enforce conservation laws across subdomains. By imposing continuity constraints at subdomain interfaces, cPINNs improve physical consistency while using domain decomposition for efficient training. An extension of this approach to generic PDEs and arbitrary space-time domains is extended PINNs (XPINNs) \cite{jagtap2020xpinns}, which partition the computational domain into subdomains and assign independent neural networks to each region, allowing parallel training. Variational approaches such as hp-variational PINNs (hp-VPINNs) instead enforce local weak-form residuals on subdomains \cite{Kharazmi2021hpvpinns}. Self-adaptive PINNs with time-domain decomposition (SA-PINN-TDD) combine adaptive loss weighting with uniform or nonuniform time partitions to improve training over long time intervals \cite{hu2026sapinnstdd}.

Another approach is to use finite-basis physics-informed neural networks (FBPINNs) \cite{moseley2023fbpinns}, which localize the approximation of the solution via overlapping subdomains and window functions that form a partition-of-unity (POU). The overlapping windowed ansatz yields a continuous global approximation without explicit interface penalties. Inspired by multilevel Schwarz methods, multilevel FBPINNs incorporate several levels of domain decomposition into the solution ansatz and have improved performance over standard PINNs on high-frequency and multiscale benchmarks \cite{dolean2024mlfbpinn}. FBPINNs have also been used in the context of extreme learning machines (ELMs) and random feature methods (RFMs) \cite{huang2006elm, rahimi2007randomfeatures, chen2022rfm, anderson2026elmfbpinns}, with related Schwarz preconditioners for randomized networks \cite{shang2025randomizedschwarz}. FBPINN training has been studied through Schwarz-based additive, multiplicative, and hybrid strategies \cite{dolean2024fbpinnschwarz}, Gauss--Newton training \cite{heinlein2025gaussnewton}, and multi-preconditioned limited-memory Broyden--Fletcher--Goldfarb--Shanno (L-BFGS) optimization using subdomain-local quasi-Newton corrections \cite{salvado2026mplbfgs}. Further developments include multifidelity FBPINNs and physics-informed neural operators for time-dependent problems \cite{heinlein2024multifidelity}, as well as model-parameter discovery in systems of ordinary differential equations (ODEs) \cite{saha2025modeldiscovery}. The same localization principle also extends to Kolmogorov--Arnold networks (KANs) \cite{howard2025fbkans}.

Related approaches adapt the partition through spatial clustering and dynamic graph-based adjustment in adaptive domain decomposition physics-informed neural networks (ADD-PINN) \cite{chen2026addpinn}, Luo et al.'s progressive domain decomposition, which uses residual-loss dynamics to guide segmentation and subdomain training \cite{luo2025progressivedd}, or a trainable gating network that defines a soft partition in augmented physics-informed neural networks (APINNs) \cite{hu2023apinns}. Ka et al.'s two-stage approach uses the residual profile of a global PINN to place subdomain boundaries for traffic state estimation \cite{ka2026adaptive}. Adaptive-basis physics-informed neural networks (AB-PINNs) learn radial-basis subdomains that move and deform during training; their AB-PINN+ variant adds subdomains in high-residual regions \cite{botvinick2026abpinns}. PINN Balls learns a local mixture-of-experts decomposition through adversarial adaptive sampling \cite{bonfanti2026pinn, tang2023aas}. For inverse problems with unknown material interfaces, Tang et al. learn a geometric partition using physics feedback and sparse observations \cite{tang2026adaptiveinterfaces}.

Other approaches adapt training or interface coupling within a domain decomposition rather than the partition geometry. These include the dual adaptive domain decomposition physics-informed neural network (DADD-PINN), which combines adaptive sampling and dynamic loss weighting with inter-subdomain feature coordination \cite{xiong2026dadd}, and Li et al.'s collaborative physics-informed neural network (CoPINN), which adapts interface conditions and loss scaling \cite{li2026copinn}. Wang et al. combine Ritz-PINN subdomain solves with residual-adaptive dataset updates for elliptic variational inequalities \cite{wang2026variationaldd}, while Chen et al. apply XPINNs to one-dimensional layered-soil consolidation, incorporating adaptive activation functions and residual-based adaptive resampling \cite{chen2026consolidationdd}.

In this work, we propose adaptive domain decomposition finite-basis physics-informed neural networks (ADD-FBPINNs). We combine partition-of-unity networks (POUNets, \cite{lee2021partitionunitynetworksdeep}) with FBPINNs to learn the number and location of subdomains. POUNets \cite{lee2021partitionunitynetworksdeep} are a data-driven approach inspired by the partition-of-unity approach to function approximation, where the solution is represented as a combination of local polynomial basis functions. POUNets simultaneously learn the partitions of the computational domain and the coefficients of the basis functions on each subdomain. We use POUNets in this work to learn window functions that define subdomains for FBPINN training. POUNets learn irregularly shaped partitions of the problem domain. However, we observe that, in the cases we study, regularly shaped partitions result in more accurate FBPINN predictions. We therefore propose separable POUNets, which use a per-axis softmax to generate more regularly shaped domains. We also introduce a Laplacian penalty on the window functions to promote sufficient subdomain overlap. POUNets have successfully been extended to multilevel domain decompositions to accelerate adaptive partitioning \cite{trask2022hierarchical}, and in \cite{rodriguez2024partition}, a POUNet is used to learn a domain decomposition for the forcing data that is then used to train a PINN.

The rest of this paper is structured as follows. In Section \ref{sec:methods}, we present our methodology, including mathematical formulations of PINNs, FBPINNs, and POUNets, followed by our ADD-FBPINN framework. Section \ref{sec:results} contains the problems of interest, an ablation study on the overlap width of subdomains in FBPINNs, and the numerical results of our ADD-FBPINN. We conclude with a discussion of the implications of these results as well as directions for further work in Section \ref{sec:disc}.

\section{Methodology}\label{sec:methods}

\subsection{Physics-informed neural networks}\label{sec:pinns}

PINNs approximate the solution to a PDE by minimizing a loss function that enforces a differential operator given by the governing equations. For simplicity, we present scalar PDEs; the framework extends to vector-valued systems without changing the domain decomposition procedure, using componentwise constraints and squared Euclidean norms of the residuals. We define a scalar PDE over an open, bounded domain $\Omega \subset \mathbb{R}^{d}$ with boundary $\partial\Omega$ as follows:
\begin{equation}
    \begin{aligned}
        \mathcal{D}[u]   & =f,   & \*x\in \Omega, \quad t\in [0,T],                           \\
        \mathcal{B}_k[u] & =g_k, & \*x\in \Gamma_k,\quad k= 1,2,\ldots, K, \quad  t\in [0,T], \\
        \mathcal{G}[u]   & = h,  & \*x\in\Omega, \quad t=0.
    \end{aligned}
\end{equation}
where $\*x$ is a $d$-dimensional input vector in $\Omega\subset\mathbb{R}^d,$ $\partial\Omega$ denotes the boundary of the domain, $\mathcal{D}$ represents the differential operator, $\mathcal{B}$ denotes the operator enforcing the boundary conditions and $\mathcal{G}$ denotes the operator that enforces initial conditions. The solution to the PDE is denoted by $u:\Omega\times[0,T]\to\mathbb{R},$ with forcing term $f,$ initial condition $h,$ and boundary conditions $g_k.$ The total number of boundary conditions is $K$ and $\Gamma_k$ represents the $k$-th boundary, so that $\bigcup_{k=1}^K \Gamma_k = \partial \Omega$. The trainable parameters (weights and biases) for the network are denoted by $\theta$, which are updated to minimize the following loss function:
\begin{equation}
    \mathcal{L}(\theta)=\lambda_{ic}\mathcal{L}_{ic}(\theta)+\lambda_{bc}\mathcal{L}_{bc}(\theta)+\lambda_r\mathcal{L}_r(\theta), \label{eq:loss}
\end{equation}
with subscripts $ic, bc, r$ corresponding to loss terms with respect to initial conditions, boundary conditions, and PDE residual. We denote the neural network by $\mathcal{NN}$ and sample $N_{ic}$, $N_{bc}$, and $N_r$ points in the space-time domain for the initial conditions, boundary conditions, and residual collocation points, respectively. The training collocation points are denoted by $\{(\*x_{ic}^i, 0)\}_{i=1}^{N_{ic}}, \{(\*x_{k_i}, t_{k_{i}})\}_{i=1}^{N_{bc}},$ and $\{(\*x_r^i, t_r^i)\}_{i=1}^{N_r}$. A common choice is to define each loss term as the mean squared residual:
\begin{align}
    \mathcal{L}_{ic}(\theta) & = \frac{1}{N_{ic}} \sum_{i = 1}^{N_{ic}} \left[ \mathcal{G}[\mathcal{NN}(\*x_{ic}^i, 0; \theta)] - h(\*x_{ic}^i, 0) \right]^2,         \\
    \mathcal{L}_{bc}(\theta) & = \sum_{k=1}^K\frac{1}{N_{bc}} \sum_{i = 1}^{N_{bc}} \left[ \mathcal{B}_k[\mathcal{NN}(\*x_{k_i}, t_{k_i}; \theta)] - g_k(\*x_{k_i}, t_{k_i})\right]^2, \\
    \mathcal{L}_{r}(\theta)  & = \frac{1}{N_{r}} \sum_{i = 1}^{N_{r}} \left[ \mathcal{D}[\mathcal{NN}(\*x_r^i, t_r^i; \theta)]-f(\*x_r^i, t_r^i)\right]^2,
\end{align}
where $\mathcal{NN}(\*x, t; \theta)$ denotes the output of the network with parameters $\theta$ at $(\*x, t)$.

The multi-objective loss function (Equation \ref{eq:loss}) must balance the minimization of the residual, boundary, and initial condition terms, which can contribute to the failure of PINNs \cite{wang2020fail}. As such, the selection of loss weights $\lambda_{ic}, \lambda_{bc}, \lambda_r$ significantly impacts how well a PINN approximates the solution to the PDE, as the weights help to ensure the loss terms are well balanced. Many methods for adaptively selecting and updating these weights have been proposed and used in practice, including soft-attention mechanism weights \cite{mcclenny2023} and neural tangent kernel weights \cite{wang2020fail}. Further, the self-adaptive hybrid weighting and sampling scheme presented in \cite{chen2025self} suggests that adaptive re-weighting is most successful when coupled with adaptive resampling. In this work, when a multi-objective loss function is required, we fix the loss weights as $\lambda_ic = 1, \lambda_{bc}= 100, \lambda_r = 1$ to highlight the efficacy of adaptive domain decomposition. However, in the following Section \ref{sec:strong}, we present an alternative method for hard constraining the boundary and initial conditions, and use this approach when possible.

\subsubsection{Hard constrained PINNs}\label{sec:strong}

The multi-objective loss function present in PINNs can cause complications during gradient descent. Another approach, presented in \cite{lagaris1998}, is to develop a solution ansatz that strictly enforces the boundary or initial conditions, or both. More precisely, the solution to the PDE is approximated by $\hat{u}(\*x;\theta)=\mathcal{C}[\mathcal{NN}(\*x;\theta)]\approx u(\*x)$ where $\mathcal{C}$ is a constraining operator that exactly enforces the boundary or initial condition. We use this approach when such an ansatz can be easily defined.

For instance, in this work, we focus on example problems with Dirichlet boundary conditions, given by
\begin{equation}
    u(\*x)=g_D(\*x), \quad \*x\in\partial\Omega,
\end{equation}
Here, $g_D:\overline{\Omega}\to\mathbb{R}$ is a sufficiently smooth function whose restriction to $\partial\Omega$ matches the prescribed Dirichlet data.
To exactly impose these conditions, we can choose a smooth scalar function $\phi_D:\overline{\Omega}\to\mathbb{R}$ with $\phi_D>0$ in $\Omega$ and $\phi_D\vert_{\partial\Omega}=0,$ and set
\begin{equation}
    \hat{u}_D(\*x;\theta)=g_D(\*x)+\phi_D(\*x)\mathcal{NN}(\*x;\theta).
\end{equation}
Then $\hat{u}_D\vert_{\partial\Omega}=g_D\vert_{\partial\Omega}$ holds by construction, independently of $\theta.$ The choice of $\phi$ is problem dependent, and we document the choice of $\phi$ for each problem in the results section. When all boundary and initial conditions are imposed exactly, training only requires minimizing the interior residual of the governing PDE. Loss terms are retained for any conditions not imposed exactly.

Note that this specific approach is restricted to problems with Dirichlet boundary conditions. However, we also test our method on problems with periodic boundary conditions, where hard constraining boundary terms is more challenging for FBPINNs. For instance, to enforce periodic boundary conditions, one can apply a Fourier embedding \cite{dong2021, wang2022respectingcausalityneedtraining} inside each local network $\mathcal{NN}_i, $ but the window function applied to each network may break periodicity. So, in this work, we strongly enforce boundary/initial conditions for problems with Dirichlet boundary conditions, whereas for the periodic Burgers example, the initial condition is imposed exactly through the solution ansatz and periodic boundary conditions are included as penalty terms in the loss alongside the PDE residual, with fixed weights as described in Section~\ref{sec:pinns}.

\subsection{Finite-basis PINNs}

Finite-basis PINNs (FBPINNs) \cite{moseley2023fbpinns} use many small neural networks on overlapping subdomains whose outputs are summed to represent the solution on the whole problem domain. They have shown superior performance in cases where a solution exhibits high-frequency or multiscale features.

Given a problem domain $\Omega\subset \mathbb{R}^d,$ we divide $\Omega$ into $L$ overlapping subdomains $\Omega_i.$ Each subdomain is assigned a separate neural network $\mathcal{NN}_i(\mathbf{x}; \mathbf{\theta}_i),$ and smooth, differentiable window functions $w_i(\mathbf{x})$ are used to combine their outputs into a global solution. The approximate solution $\hat{u}(\*x; \*\theta)$ is given by
\begin{equation}
    \hat{u}(\*x; \*\theta) = \mathcal{C}[\overline{\mathcal{NN}}(\*x;\theta)],
\end{equation}
where
\begin{equation}\label{eq:fbpinn_input}
    \overline{\mathcal{NN}}(\*x;\theta) = \sum_{i=1}^L w_i(\*x)\cdot\mathcal{NN}_i(\text{norm}_i(\*x); \theta_i).
\end{equation}
Here, $\mathcal{NN}_i(\text{norm}_i(\*x); \theta_i)$ is the neural network applied to domain $\Omega_i,$ $\text{norm}_i(\*x)$ denotes the normalization applied to inputs of the subdomain, $\mathcal{C}$ is a constraining operator that enforces boundary or initial conditions on the solution ansatz (as in Section \ref{sec:strong}), and $w_i(\*x)$ is a smooth scalar window function that confines each network to its subdomain. We train the FBPINNs in this work using the ADAM optimizer. In particular, these window functions ensure each neural network contributes only within its respective subdomain, {with its contribution identically zero outside that region}. This means that the neural network only has to be evaluated on collocation points within its respective subdomain ${\*x}\in\Omega_i.$ The window functions satisfy a POU constructed with the expression
\begin{equation}
    w_i=\frac{\hat{w}_i}{\sum_{{j}=1}^L\hat{w}_{{j}}} \quad \text{where} \quad \text{supp}({\hat{w}_i})\subset\overline{\Omega_i},
\end{equation}
where $\hat{w}_i$ denotes the unnormalized window function. Here, $L$ represents the total number of subdomains. The unnormalized window functions can be, for instance, chosen as:
\begin{equation}\label{eq:fbpinn_windows}
    {\hat{w}_i}(\*x)=\prod_{j=1}^d\psi\left(\frac{x^j-{c_i^j}}{h^j_i}\right), \quad \text{where} \quad \psi(t)=\begin{cases}
        (1+\cos(\pi t))^2, & \vert t\vert\leq1, \\
        0,                 & \text{otherwise}
    \end{cases}
\end{equation}
where ${\*c_i}\in\mathbb{R}^d$ {is the center of subdomain} $\Omega_i$ and ${\*h_i}\in\mathbb{R}^d$ {contains the half-widths of its support in each dimension}.

For uniformly spaced subdomains with center spacing $d^j = c_{i+1}^j - c_i^j$, we parameterize the half-width as
\begin{equation}\label{eq:overlap}
    h_i^j = \frac{d^j + \delta}{2},
\end{equation}
where $\delta$ denotes the total overlap width between adjacent subdomains.

The same approach used for PINNs is then used to train an FBPINN. In particular, given a set of collocation points $\{\*x_i\}_{i=1}^N,$ the loss function for the FBPINN is
\begin{equation}\label{eq:fbpinnloss}
    \mathcal{L}(\theta)=\frac{1}{N}\sum_{i=1}^N(\mathcal{D}\hat{u}(\*x_i;\theta)-{f}(\*x_i))^2
\end{equation}
\subsection{Partition-of-Unity Networks}

Partition-of-Unity Networks (POUNets), presented in {\cite{lee2021partitionunitynetworksdeep, trask2022hierarchical}}, are used to approximate functions by learning POU window functions that partition the problem domain with a neural network while polynomial spaces with learnable coefficients make higher-order approximations on each partition. For our approach, we use the POUNets in an iterative manner to learn the window functions that are later used as subdomains for training an FBPINN, as detailed in Section \ref{sec:adaptiveDD}. In this section, we present POUNets as they pertain to our method.

Consider a POU $\Phi=\{\omega_{i}(\*x)\}_{i=1}^{N_{\text{part}}}$ satisfying $\sum_{i}\omega_{i}(\*x)=1$ and $\omega_{i}(\*x)\geq0$ for all $\*x.$ POUNets approximate {a scalar function} $y$ with the approximant
\begin{equation}
    y_{{pou}}(\*x)=\sum_{{i}=1}^{N_{\text{part}}}\omega_{i}(\*x)\sum_{j=1}^{\dim(V)}c_{i,j}P_{j}(\*x)
\end{equation}
where $V=\text{span}\{P_{j}\}$ and $c_{i,j}\in\mathbb{R}$ are coefficients in the polynomial approximations. A common basis choice is $V=\pi_m(\mathbb{R}^d),$ which is the space of polynomials of order $m$ {in $d$ variables}.

One approach is to construct the POU $\Phi$ prior to fitting the coefficients $c_{i,j}$ through geometric meshing. Instead, POUNets represent the POU as a neural network, in which the weights and biases $\xi$ are trained to fit the data in the $L_2$ sense. A variety of architectures can be used to represent these window functions. The original work \cite{lee2021partitionunitynetworksdeep} {used a shallow radial basis function (RBF) network and a standard multilayer perceptron (MLP)} to represent the window functions. We relegate the formulation of the MLP architecture to Appendix \ref{app:pou_arch}, but note that the learnable parameters $\theta_{\text{pou}}$ for the MLP are the weights and biases of the neural network. The RBF-Net \cite{govindaraju2000RBF, billings1995RBF} has the form
\begin{equation}\label{eq:RBF}
    \omega_{i}(\*x;\theta_{\text{pou}})=\frac{\exp(-\vert\*x-\*\xi_{1,i}\vert^2/\xi_{2,i}^2)}{\sum_{j}\exp(-\vert\*x-\*\xi_{1,j}\vert^2/\xi_{2,j}^2)}, \quad 1\leq i\leq N_{\text{part}}.
\end{equation}
This shallow RBF-network uses a single hidden layer without additional nonlinear transformations. Here, the learnable parameters for the POUNet are $\theta_{\text{pou}}=(\*\xi_1, \xi_2),$ where $\*\xi_1$ are the RBF centers and $\xi_2$ are the RBF shape parameters. The output of the RBF-network is determined by the centers and width parameters of the RBF functions, without additional nonlinear layers.

\begin{figure}[htbp]
    \centering
    \includegraphics[width=0.7\linewidth]{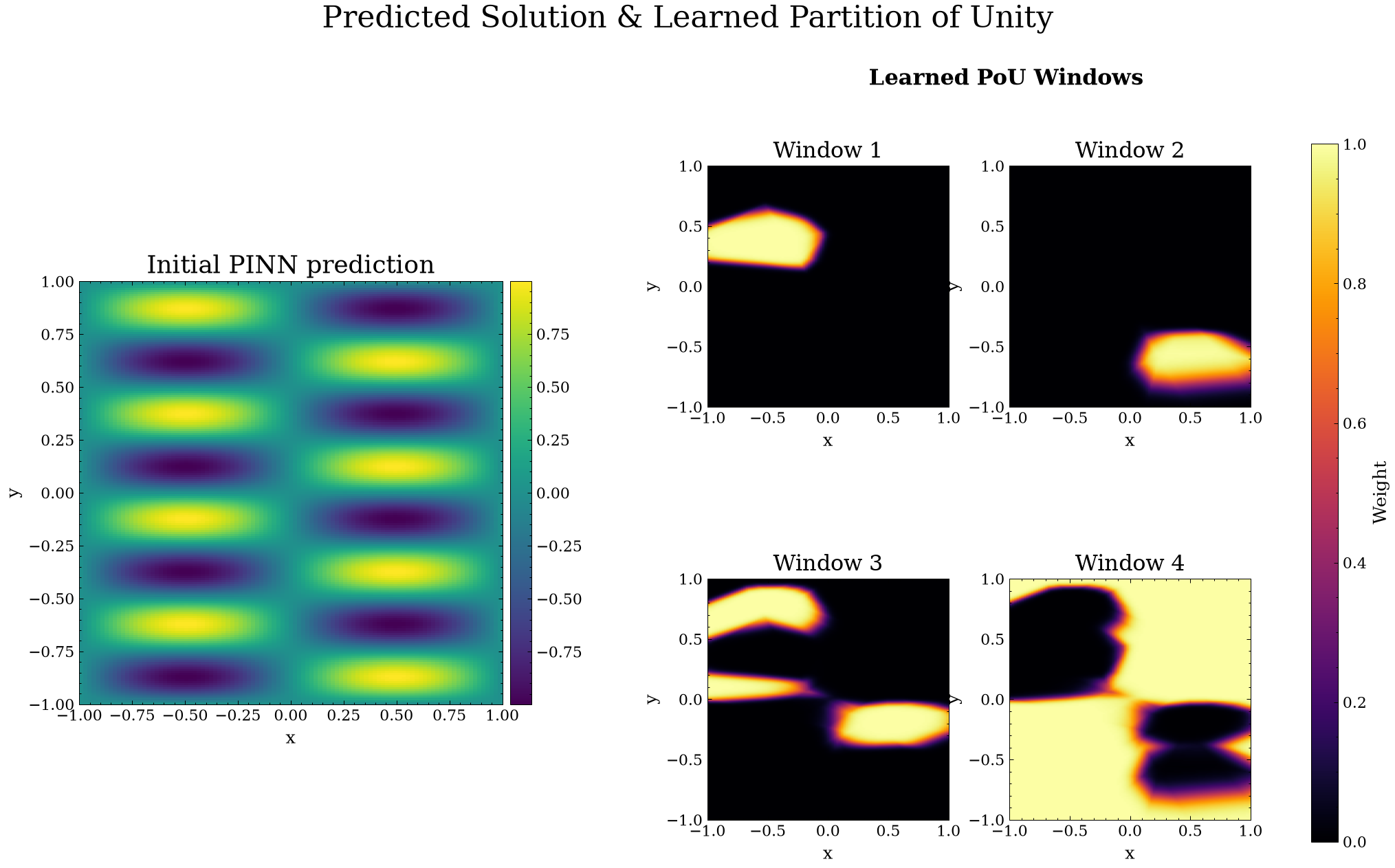}
    \caption{Predicted solution by the initial PINN for the Helmholtz equation (Eq. \ref{eq:Helmholtzeq}) with $a_1=1, a_2=4$ (left) and learned subdomains by the {POUNet} (right). In this figure, the prediction on the left is used as data for the {POUNet}, which outputs the subdomains on the right. }
    \label{fig:mlp_domains}
\end{figure}

In practice, the partitions learned by a POUNet with either the MLP or RBF architecture are often irregular in shape. An example of this phenomenon can be seen in Fig. \ref{fig:mlp_domains}, where the subdomains group according to the regions of high-frequency. Depending on the problem of interest, however, the FBPINN may perform better with a domain decomposition that consists of regular-shaped partitions. And so, we also present Separable POUNets as an alternative POUNet architecture in Section \ref{sec:sep} that generates more regular partitions.

The localized basis coefficients and the localization (POU) are fit simultaneously by minimizing a dual objective loss function:
\begin{equation}\label{eq:POUobj}
    \min_{\*\theta_{pou}, \*c}\sum_{i=1}^{N_{data}}\left\vert\sum_{\alpha=1}^{N_{part}}\phi_{\alpha}(\*x_i;\*\theta_{pou})\sum_{\beta=1}^{\dim(V)}c_{\alpha,\beta}P_{\beta}(\*x_i)-y_i\right\vert^2.
\end{equation}
{Here, $N_{data}$ is the number of training samples, $N_{part}$ is the number of partitions, and $\phi_\alpha=\omega_\alpha$ denotes a window function.} We use least-squares gradient descent (LSGD) \cite{cyr2019robusttraininginitializationdeep} to alternate between updating the polynomial approximation parameters $\*c$ and the network parameters for the network that represents the window function $\*\theta.$ We describe this process in Algorithm \ref{alg:LSGD}. In essence, we first fix $\theta_{pou},$ and update $\*c$, which has a closed-form solution because it is {a typical least-squares (LS) problem}. Then, fix $\*c^*,$ and {take one gradient descent (GD) step} to update $\*\theta$ with the loss function
\begin{equation}\label{eq:POUloss}
    \mathcal{L}_{pou}(\*\theta_{pou}, \*c^*) = \sum_{i=1}^{N_{data}}\vert y_{pou}(\*x_i; {\*\theta_{pou}}, \*c^*)-y_i\vert^2.
\end{equation}
In our work, we introduce the addition of an optional penalty term to the loss function of the POUNet to enforce wider overlaps. POUNets are trained to minimize an objective that rewards localized window functions, which in practice results in sharp transitions between subdomains. However, FBPINNs require sufficient overlap width between subdomains to ensure continuity of the learned solution. Our solution to this problem is detailed in Section \ref{sec:penalty} and is further motivated by our fixed domain decomposition ablation studies in Section \ref{sec:ablation}.

\begin{algorithm}
    \caption{Least-squares Gradient Descent (LSGD) \cite{cyr2019robusttraininginitializationdeep}}  \label{alg:LSGD}
    \begin{algorithmic}[1]
        \Input $\theta_{\text{pou, old}}, \*c_{\text{old}}, n_{\text{epochs}}, \eta$
        \Output $\theta_{\text{pou, new}}, \*c_{\text{new}}$
        \Function{LSGD}{$\theta_{\text{pou, old}}, \mathbf{c}_{\text{old}},
n_{\text{epochs}},  \eta$}
    \State $\theta_{\text{pou}} \leftarrow \theta_{\text{pou, old}}$
    \For{$k=1,\ldots,n_{\text{epochs}}$}
        \State $\mathbf{c} \leftarrow \operatorname{LS}(\theta_{\text{pou}})$
        \State $\theta_{\text{pou}} \leftarrow
        \operatorname{GD}(\theta_{\text{pou}}, \mathbf{c}, \eta)$
    \EndFor
    \State $\mathbf{c}_{\text{new}} \leftarrow
    \operatorname{LS}(\theta_{\text{pou}})$
    \State $\theta_{\text{pou,new}} \leftarrow \theta_{\text{pou}}$
    \State \Return $\theta_{\text{pou,new}}, \mathbf{c}_{\text{new}}$
\EndFunction
    \end{algorithmic}
\end{algorithm}

\subsubsection{Separable {POUNets}}\label{sec:sep}

For the problems we consider, regular or nearly-regular subdomains consistently produce the most accurate FBPINN approximation due to symmetry. As such, we propose separable {POUNets}, which aim to generate more regular partitions by performing a per-axis softmax. While this approach generalizes to ${d}$ dimensions, {we present the approach in two dimensions (2D).}

For $\*x=(x_1,x_2)\in\mathbb{R}^2$ and total number of partitions $N=n_{x_1}\times n_{x_2}$, where $n_{x_1}$ and $n_{x_2}$ are the number of partitions along the $x_1$ and $x_2$ dimensions, respectively, each partition, indexed by $(i,j),$ has an associated partition function:
\begin{equation*}
    \{w_{ij}(\*x)\}_{i=1, j=1}^{n_{x_1}, n_{x_2}}.
\end{equation*}
Two independent one-dimensional functions $f_{x_1}, f_{x_2},$ in our case MLPs or RBF-nets, produce axis-wise predictions:
\begin{equation*}
    w_{ij}(\*x)= \frac{\exp(f_{x_1}(x_1;\theta_{x_1})_i+f_{x_2}(x_2;\theta_{x_2})_j)}{\sum_{p=1}^{n_{x_1}}\sum_{q=1}^{n_{x_2}}\exp(f_{x_1}(x_1;\theta_{x_1})_p+f_{x_2}(x_2;\theta_{x_2})_q)}, \quad i=1,\ldots,n_{x_1}, j=1,\ldots, n_{x_2}.
\end{equation*}
The denominator factorizes,
\begin{equation*}
    \sum_{p=1}^{n_{x_1}}\sum_{q=1}^{n_{x_2}}\exp(f_{x_1}(x_1;\theta_{x_1})_p+f_{x_2}(x_2;\theta_{x_2})_q) = \left(\sum_{p=1}^{n_{x_1}}\exp(f_{x_1}(x_1;\theta_{x_1})_p)\right)\left(\sum_{q=1}^{n_{x_2}}\exp(f_{x_2}(x_2{;}\theta_{x_2})_q)\right),
\end{equation*}
and we can rewrite $w_{ij}$ in the fully factorized form:
\begin{equation}
    w_{ij}(\*x)=\frac{\exp(f_{x_1}(x_1;\theta_{x_1})_i{)}}{\sum_{p=1}^{n_{x_1}}\exp(f_{x_1}(x_1;\theta_{x_1})_p)}\frac{\exp(f_{x_2}(x_2;\theta_{x_2})_j{)}}{\sum_{q=1}^{n_{x_2}}\exp(f_{x_2}(x_2{;}\theta_{x_2})_q)}.
\end{equation}
These partitions satisfy the POU condition, ensuring that their sum is always one at any point in the domain: $\sum_{i=1}^{n_{x_1}}\sum_{{j}=1}^{n_{x_2}}w_{ij}(\*x)=1\quad \forall \*x\in\Omega.$

And so, the single softmax is algebraically identical to first normalizing each axis ($\sum_iw_i=1, \sum_jw_j=1$) and then taking their outer product. This compact formulation preserves the POU property. We test both Separable-MLP-{POUNets} and Separable-RBF-{POUNets} alongside the standard MLP and RBF-net formulations in this work, but our default setting for each problem is the Separable-RBF-{POUNet} because the subdomains derived from the Separable-RBF-{POUNets} {often result in better} final error approximations after FBPINN training.

\subsubsection{Overlap regularization term}\label{sec:penalty}

We also introduce an overlap regularization term to the loss function of the POUNet to ensure that the learned domain decomposition includes wide enough overlapping regions between subdomains. The necessity of this is motivated in our ablation studies in Section \ref{sec:ablation}, in which we show that simply adding collocation points to the overlapping regions cannot reliably improve accuracy without increasing overlap width.

We introduce a penalty term to Equation \ref{eq:POUobj} that penalizes the Laplacian of the window functions. The Laplacian of a window function $w_{ij}$ controls the curvature of the window function, so this penalty term penalizes rapid blending between subdomains. 
The penalty is given by
\begin{equation}\label{eq:lap_loss}
    \mathcal{L}_{lap}(\*\theta_{pou})=\frac{1}{N}\sum_{n=1}^N{\sum_{i=1}^{n_{x_1}}\sum_{j=1}^{n_{x_2}}\left(\Delta w_{ij}(\*x_n;\theta_{\text{pou}})\right)^2}
\end{equation}

\begin{figure}[htbp]
    \centering
    \includegraphics[width=0.9\linewidth]{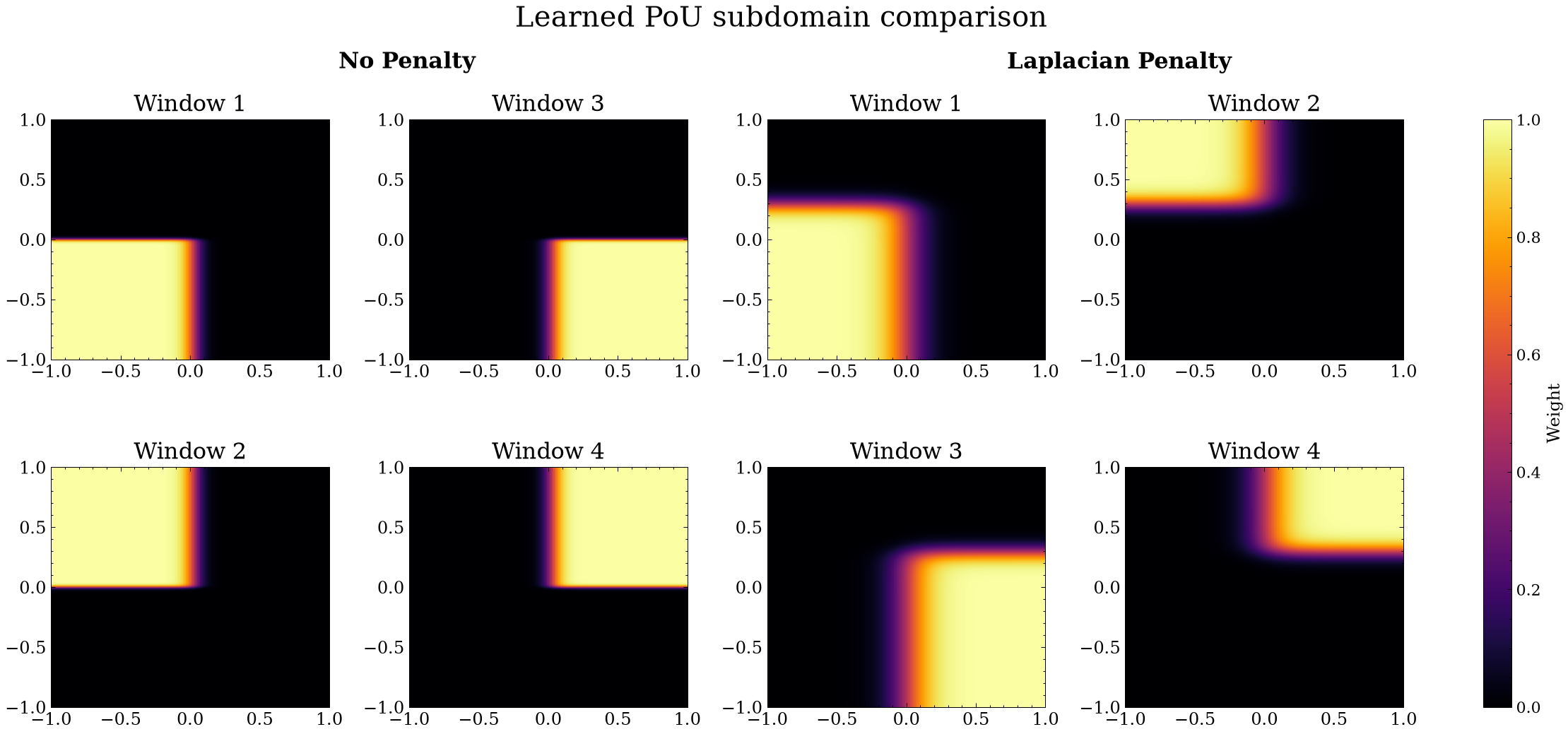}
    \caption{Comparison of the subdomains learned by a {POUNet} for the Helmholtz equation with no added penalty terms to enforce wider overlaps (left), and the subdomains learned by a {POUNet} with Laplacian penalty term (right). The Laplacian penalty term has a weight of $\lambda_{lap}=0.0001.$ The subdomains with Laplacian penalty have a wider transition between subdomains, which aids in downstream FBPINN training.}
    \label{fig:penalty_comparison}
\end{figure}

{Here, $N$ is the number of points used to evaluate the penalty.} With the Laplacian penalty term, our composite loss function for a POUNet becomes
\begin{equation}
    \mathcal{L}(\*\theta_{pou}, \*c^*) = \mathcal{L}_{pou}(\*\theta_{pou}, \*c^*)+\lambda_{lap}\mathcal{L}_{lap}(\*\theta_{pou})
\end{equation}
For our experiments, we fix the weight on the Laplacian penalty term to be $\lambda_{\text{lap}}=0.001.$ To demonstrate the efficacy of this added penalty term, we use a POUNet to learn a subdivision of the problem domain for the {2D} Helmholtz problem presented in Equation \ref{eq:Helmholtzeq}. To highlight the impact of the overlap term,  we plot the difference between the window functions learned from a POUNet with and without the overlap term in Fig. \ref{fig:penalty_comparison}. In practice, no overlap penalization results in {very sharp} partition boundaries, which represents successful POUNet training as developed in \cite{lee2021partitionunitynetworksdeep}, but is ill-suited for downstream FBPINN training.

\subsection{ADD-FBPINNs}\label{sec:adaptiveDD}

We propose adaptive domain decomposition via a hierarchical approach that relies on a {POUNet} to learn {subdomains} on which an FBPINN is trained. The integration of {POUNets} and FBPINNs is motivated by the need for flexible, physics-informed domain decomposition for solving PDEs. A key benefit of our method is that {no additional data are needed for learning the domain decomposition}.  Since {POUNets} are a data-driven approach, we use a naive approximation given by a standard PINN to facilitate the {POUNet} training. {POUNets} allow the neural networks to focus on partitioning the problem domain, and we utilize the learned partitions from a {POUNet} to define subdomains on which to train an FBPINN.

\begin{figure}[htbp]
    \centering
    \includegraphics[width=0.9\linewidth]{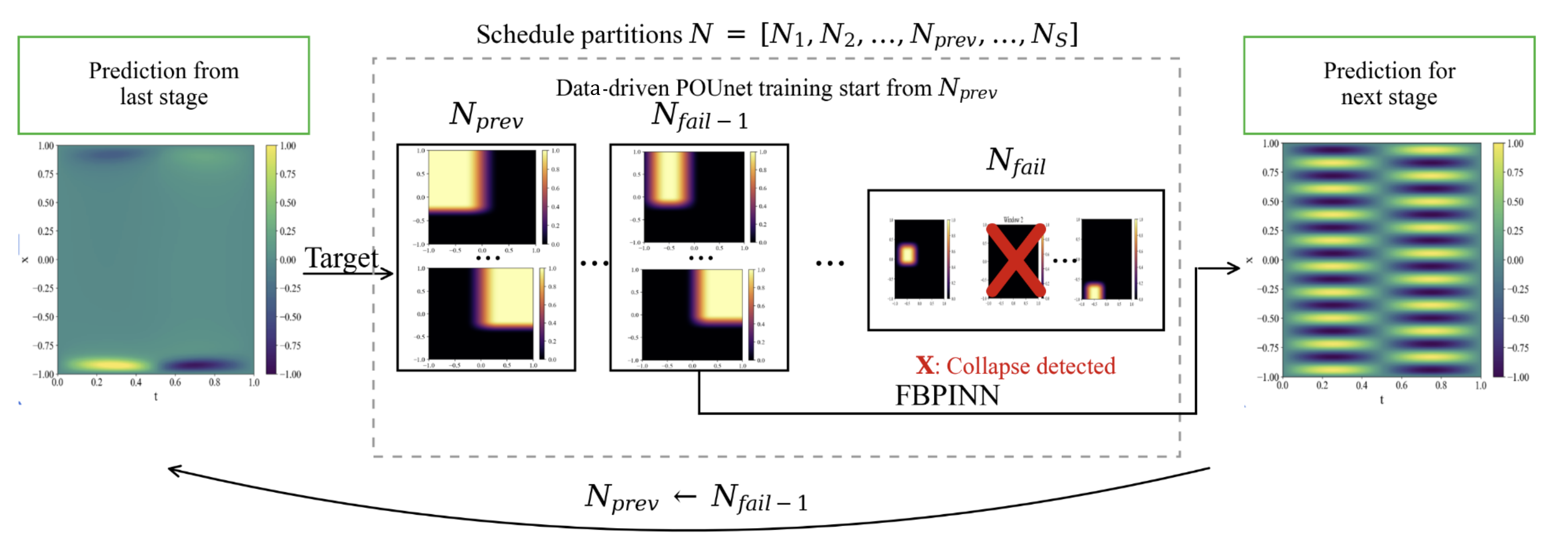}
    \caption{Graphical abstract of the ADD-FBPINN}
    \label{fig:graphical_abs}
\end{figure}

We present our approach in Algorithm \ref{alg:adaptiveDD} and provide an illustration in Fig. \ref{fig:graphical_abs}. This adaptive algorithm has the following stages:

\begin{enumerate}
    \item \textbf{Initialization}: A baseline PINN is trained on the entire problem domain $\Omega$ to generate a low-fidelity approximation ${\hat{u}}^{(1)}.$ This solution serves as the first target, $\*y_{\text{target}},$ {a vector of sampled scalar solution values,} for the {POUNet}. A schedule of increasing number of partitions $\{N_1, N_2, \ldots, N_S\}$ is defined.
    \item \textbf{Iterative partitioning and refinement}: The algorithm iterates through the partition schedule. For each number of partitions $N=N_1, N_2, \ldots, N_S$:
          \begin{enumerate}[label=\alph*.]
              \item \textbf{{POUNet} training:} A {POUNet} is trained to learn window functions for a set of $N$ partitions, $\Phi^{(N)},$ using the most recent prediction $\*y_{\text{target}}.$
              \item \textbf{Collapse detection:} The algorithm checks if any partition has ``collapsed,'' i.e., at least one window function has maximum value drop below the threshold $\tau.$
                    \begin{itemize}
                        \item \textbf{If a collapse occurs:} The current partition number is marked as $N_{\text{fail}}.$ The algorithm reverts back to the last successful partition configuration ($\Phi^*$ at $N_{\text{prev}}$) and trains an FBPINN using the window functions $\Phi^*$ as subdomains to refine the approximate solution to the PDE. {The new} solution becomes the updated $\*y_{\text{target}},$ and the {POUNet} is retried on $N_{\text{fail}}.$ If the collapse occurs again, training terminates and returns the last successful result.
                        \item \textbf{If no collapse occurs:} The learned partitions $\Phi^{(N)}$ are accepted as the new best configuration $\Phi^*,$ and the algorithm proceeds to the {next partition} number in the schedule.
                    \end{itemize}
          \end{enumerate}
    \item \textbf{Final FBPINN training:} Once the loop completes, either by exhausting the schedule or terminating early due to a repeated collapse, a final FBPINN is trained on the last successful number of partitions $\Phi^*.$ This yields the final solution.
\end{enumerate}

\begin{algorithm}[htbp]
    \caption{ADD-FBPINN}\label{alg:adaptiveDD}
    \begin{algorithmic}[1]
        \item Train PINN and obtain $\hat{u}^{(1)}$; \quad $\*y_{\text{target}}\leftarrow\hat{u}^{(1)}$
        \item Schedule partition numbers $\{N_1, N_2, \ldots, N_S\}$; \quad choose collapse threshold $\tau$
        \item {Train POUNet at $N_1$ using $\*y_{\text{target}}$; obtain partitions $\Phi^{(N_1)} = \{\phi_i^{(N_1)}\}_{i=1}^{N_1}$}
        \item $\Phi^*\leftarrow{\Phi^{(N_1)}},$ \quad $N_{\text{prev}}\leftarrow N_1$
        \For{$N\in\{N_2, \ldots, N_S\}$}
        \State Train {POUNet} at $N$ using $\*y_{\text{target}}$; obtain partitions $\Phi^{(N)} = \{\phi_i^{(N)}\}_{i=1}^{N}$
        \If{$\exists{i}$ s.t. $\max_{\*x}\phi_i^{(N)}(\*x)<\tau$} \Comment{Collapse detected at $N$}
        \State Train FBPINN at $N_{\text{prev}}$ with $\Phi^*$ to get $\hat{u}^{(N_{\text{prev}})}$
        \State $\*y_{\text{target}}\leftarrow\hat{u}^{(N_{\text{prev}})}$ \Comment{Update target with a better solution}
        \State Train {POUNet} at $N$ using new $\*y_{\text{target}}$ and obtain new $\Phi^{(N)}$ \Comment{Retry {POUNet} training for $N$}
        \If{$\exists{i}$ s.t. $\max_{\*x}\phi_i^{(N)}(\*x)<\tau$} \Comment{Collapse again at $N$}
        \State \Return last FBPINN (at $N_{\text{prev}}$) and $\Phi^*$ \Comment{Stop}
        \EndIf
        \EndIf
        \State ${\Phi^*\leftarrow\Phi^{(N)},}$ \quad ${N_{\text{prev}}\leftarrow N}$ \Comment{Update partitions and proceed to next $N$}
        \EndFor
        \State Train FBPINN at $N_{\text{prev}}$ with $\Phi^*$
        \State \Return last FBPINN and $\Phi^*$
    \end{algorithmic}
\end{algorithm}

A partition is fully effective when its maximum value is 1, and its influence diminishes as its maximum value approaches zero, because a value near zero means the partition never achieves exclusive ownership of any point in the domain. And so, a collapse is defined as the maximum value of any partition dropping below a threshold $\tau.$ Setting $\tau=1.0$ can result in false detection of a collapse so we take $\tau=0.8$ to appropriately balance partition effectiveness and collapse detection. 

Because the POUNet windows are produced by a softmax, each window function is strictly positive on the entire domain, meaning the learned windows have global support. In practice, trained windows decay rapidly outside the region where they are close to $1$. We use this region to define an effective support to replicate the input normalization of FBPINN training (Equation \ref{eq:fbpinn_input}).

In the following section, we test this approach on the {2D} Helmholtz equation, {2D} Poisson equation with multiscale features, and viscous Burgers equation. We tested four architectures of {POUNets} in our experiments and found that the separable RBF-{POUNet} most often results in the lowest final relative $L_2$ error, and so we present only those results and discuss the impacts of the standard RBF, MLP and separable MLP {POUNet} architectures in Appendix \ref{sec:pou_comparison}.

\subsection{Sampling Methods}

The placement and spatial coverage of the collocation points in the domain $\Omega$ are critical to the training of PINNs, as they strongly affect the convergence and accuracy \cite{wu2023rad}. {For our one-dimensional (1D) example,} we default to sampling these points from a uniform distribution over $\Omega.$ For the {2D} examples, we default to the Hammersley \cite{hammersley1964} sampling approach as a low-discrepancy quasi-random sequence that provides more uniform coverage of the domain than standard random sampling.

During training, residual-based adaptive distribution (RAD) resampling \cite{wu2023rad} can be applied, which periodically resamples collocation points with probability proportional to the local PDE residual magnitude, concentrating points in regions of high error. Our experiments suggest that resampling points according to residual error results in improved FBPINN predictions. For brevity, a presentation of this approach is included in Appendix \ref{sec:rad}.

\section{Results}\label{sec:results}
\subsection{Problems studied}

\subsubsection{{1D} Sine ODE}

As an ablation test (Section \ref{sec:ablation}) on the importance of subdomain overlap width and density of collocation points in the overlapping regions, we consider the equation
\begin{equation}\label{eq:sineODE}
    \frac{du}{dx}=\sin(\omega_1x)+\sin(\omega_2x), \quad u(0)=0
\end{equation}
for $x\in [-2\pi, 2\pi]$, which has the exact solution
\begin{equation*}
    u(x)=\frac{1-\cos(\omega_1x)}{\omega_1}+\frac{1-\cos(\omega_2x)}{\omega_2}.
\end{equation*}
We consider the case where $\omega_1=1, \omega_2=45.$

\subsubsection{Helmholtz equation}
We consider the following two-dimensional Helmholtz problem. The equation in the spatial domain $(x,y)$ is given by
\begin{equation}\label{eq:Helmholtzeq}
    \begin{aligned}
         & u_{xx}+u_{yy}+k^2u-q(x,y)=0                                           \\
         & u(x,y)=0, \quad x,y\in\partial\Omega, \quad \Omega=[-1,1]\times[-1,1]
    \end{aligned}
\end{equation}
where $q(x,y)$ is the following forcing term with $k=1:$
\begin{equation*}
    \begin{aligned}
        q(x,y)= & -(a_1\pi)^2\sin(a_1\pi x)\sin(a_2\pi y)  \\
                & - (a_2\pi)^2\sin(a_1\pi x)\sin(a_2\pi y) \\
                & +k^2\sin(a_1\pi x)\sin(a_2\pi y).
    \end{aligned}
\end{equation*}
With this, we have the closed-form, analytical solution
\begin{equation*}
    u(x,y)=\sin(a_1\pi x)\sin(a_2\pi y).
\end{equation*}

For this problem, we use the solution ansatz 

\begin{equation}
    \hat{u}(x,y)= (1-x^2)(1-y^2)(NN_{\theta}(x,y))
\end{equation}

This problem is used as a benchmark test for the performance of our ADD-FBPINN method. We vary the solution mode numbers while keeping the Helmholtz wavenumber $k=1$. For this purpose, we vary the values of $a_1, a_2$ to increase the difficulty of the problem. We also use this problem with $a_1=1, a_2=4$ in an ablation study with standard FBPINNs to gauge the impact of increasing overlap width between subdomains, detailed in Section \ref{sec:ablation}.

We note that multilevel FBPINNs with a fixed domain decomposition have previously outperformed standard PINNs on a Helmholtz benchmark; see~\cite{dolean2024mlfbpinn}. However, their performance was limited at higher wavenumbers, a limitation not observed for the other scalar elliptic problems tested. In FBPINNs, windowed local contributions vanish at subdomain boundaries, while the global solution is coupled through overlap. For Helmholtz problems, optimized Schwarz methods generally yield better results than standard overlapping Schwarz methods through the use of wave-adapted transmission conditions, such as Robin conditions \cite{gander2002optimizedhelmholtz}.

\subsubsection{Poisson equation with multiscale features}

Next, we study a two-dimensional Poisson equation that is used in \cite{dolean2024mlfbpinn} given by
\begin{equation}\label{eq:Poissoneq}
    \begin{aligned}
        -\Delta u=f & \quad \text{in } \Omega=[0,1]^2, \\
        u = 0       & \quad \text{on } \partial\Omega,
    \end{aligned}
\end{equation}
where the source term $f(x_1, x_2)$ is
\begin{align}
    f(x_1, x_2) = \frac{2}{n}\sum_{i=1}^n({\omega_i}\pi)^2\sin(\omega_i\pi x_1)\sin({\omega_i}\pi x_2),
\end{align}
which corresponds to the exact solution
\begin{equation}
    u(x_1, x_2) = \frac{1}{n}\sum_{i=1}^n\sin({\omega_i}\pi x_1)\sin({\omega_i}\pi x_2).
\end{equation}
In this case, multiscale features are present in the solution, and we can tune $n$ and ${\omega_i}$ to adjust the number of components and the frequency of each component.

For this problem, we use the solution ansatz 

\begin{equation}
    \hat{u}(x,y)= x(1-x)y(1-y)(NN_{\theta}(x,y))
\end{equation}

\subsubsection{Burgers equation}

For the final benchmark, we employ two cases of the Burgers equation. Case 1 is given by
\begin{equation}
    \begin{aligned}
        \label{eq:Burgers}
         & u_t + uu_x-(0.01/\pi)u_{xx} = 0, \quad x\in[-1,1], \quad t\in[0,1] \\
         & u(0,x)=-\sin(\pi x)                                                \\
         & u(t, -1) = u(t, 1)=0
    \end{aligned}
\end{equation}
where $\nu=0.01/\pi$ is the viscosity.

We can strongly enforce boundary and initial conditions using the following ansatz:
\begin{equation*}
    u(t,x) = -\sin(\pi x)(1-t)+(1-x^2)t(NN_{\theta}(t,x))
\end{equation*}
We also consider the same PDE with a shifted initial condition, denoted Case 2,
\begin{equation}
    \begin{aligned}
        \label{eq:b_periodic}
         & u(0,x)=-\sin(\pi x+\frac{\pi}{4}) \\
         & u(t, -1) = u(t, 1)                \\
         & u_x(t,-1) = u_x(t,1).
    \end{aligned}
\end{equation}
For this example, we choose to enforce periodic boundary conditions via a term in the loss function,
\begin{equation*}
    \mathcal{L}_{bc}(\theta) = \frac{1}{N_{bc}}\sum_{i=1}^{N_{bc}}\left[\mathcal{NN}(t_i,-1,\theta) - \mathcal{NN}(t_i,1,\theta)\right]^2 + \frac{1}{N_{bc}}\sum_{i=1}^{N_{bc}}\left[\mathcal{NN}_x(t_i,-1,\theta) - \mathcal{NN}_x(t_i,1,\theta)\right]^2
\end{equation*}
but keep strongly enforced initial conditions with the ansatz:
\begin{equation*}
    \hat{u}(t,x) = -\sin(\pi x+\frac{\pi}{4})(1-t)+\frac{\sqrt{2}}{2}(1-x^2)t(NN_{\theta}(t,x))
\end{equation*}

\subsection{Impact of subdomain overlap width}\label{sec:ablation}

In this section, we perform an ablation study using standard FBPINNs with varying overlap widths. This ablation study aims first to show that increasing the overlap width can greatly improve the accuracy of FBPINNs. We demonstrate that there exist cases where an FBPINN can fail to accurately capture the dynamics of the solution, yet when the same network is trained with a larger overlap region, the prediction from the FBPINN improves significantly. To do this, we fix the number of subdomains in a non-adaptive FBPINN via Equation \ref{eq:fbpinn_windows} and vary only the overlap width. We select examples exhibiting multiscale or high-frequency behavior, for which FBPINNs have been shown to improve accuracy compared with standard PINNs \cite{moseley2023fbpinns}, to isolate the effects of varying overlap widths on FBPINN accuracy. These are the 1D Sine equation \ref{eq:sineODE}, 2D Helmholtz \ref{eq:Helmholtzeq}, and 2D Poisson \ref{eq:Poissoneq}. The 2D Helmholtz equation and 2D Poisson equation are selected as benchmarks for our ADD-FBPINN, and each equation has multiple cases tested at varying levels of frequency/multiscale features. And so, for the ablation study, we selected the lowest-frequency cases. Specific parameter choices are described below.

For each test case, we over-saturate the collocation points in the overlapping regions
with the smallest width. The extra collocation points are sampled according to the following scheme. First, we assign each overlap a density, given by
\begin{equation}
    \rho_{\text{base}}=\frac{N^d}{\vert \Omega \vert}, \quad N_{\text{extra}}=\rho_{\text{base}} \cdot \alpha \cdot \vert \Omega_{\text{overlap}}\vert.
\end{equation}
where $N$ denotes the number of collocation points in each dimension before oversampling, $\vert\Omega\vert$ denotes the size of the domain, $\vert \Omega_{\text{overlap}}\vert$ denotes the size of the overlap, and $\alpha\geq1$ is an oversampling factor that determines how many extra collocation points will be placed in the overlapping region. For this study, we set $\alpha=4.0$ to ensure that the overlapping regions are indeed oversaturated with collocation points.

\begin{figure}
    \centering
    \includegraphics[width=0.9\linewidth]{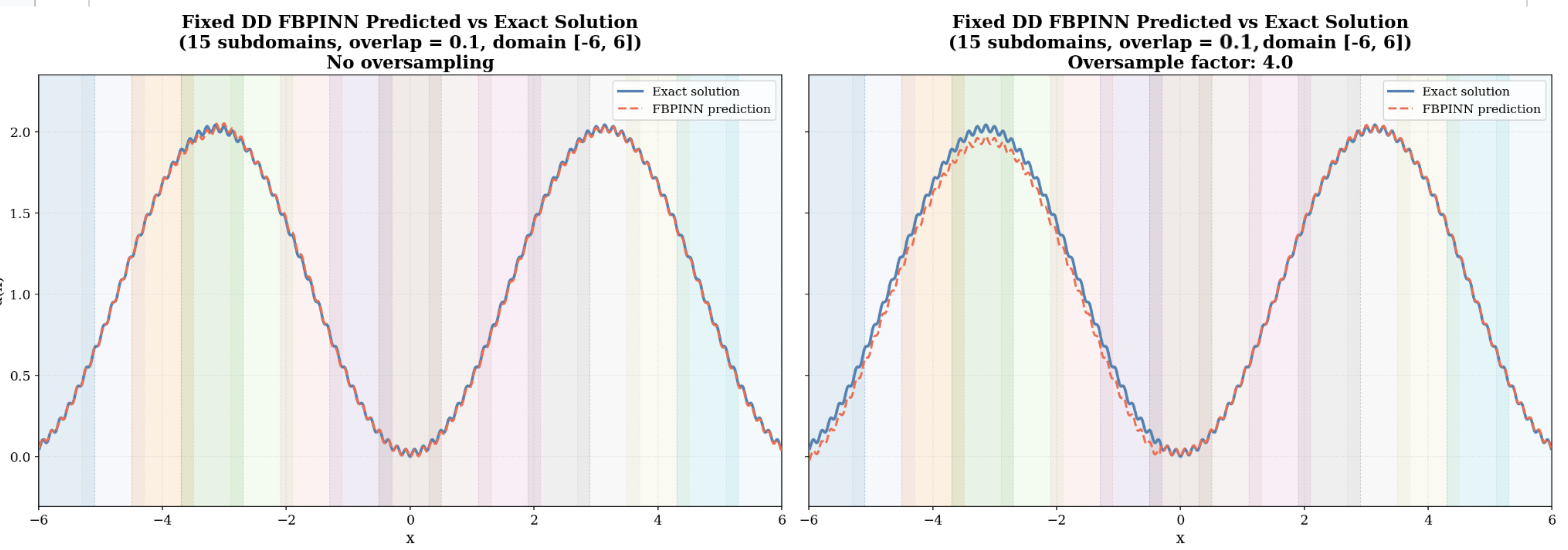}
    \caption{Predicted and exact results for the 1D Sine ODE with 2000 base points (left) and with 2000 base points + oversampling in the overlap (right).}
    \label{fig:abl_compare_oversampling}
\end{figure}

\begin{figure}
    \centering
    \includegraphics[width=0.7\linewidth]{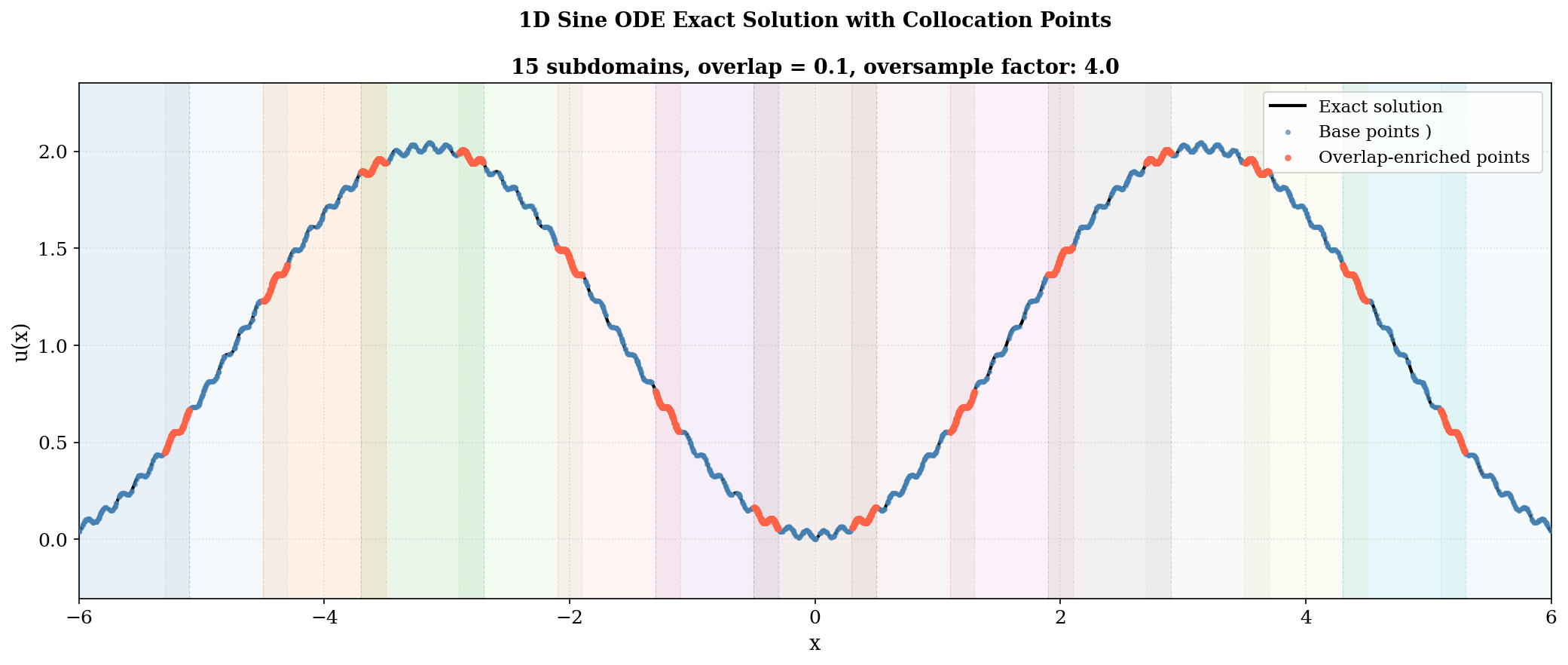}
    \caption{Collocation points used to solve the 1D Sine ODE with oversample factor of $\alpha=4.0.$ Collocation points are plotted along the exact solution. Blue points denote points in only one subdomain, and red points denote collocation points active in two subdomains.}
    \label{fig:sin1d_pts}
\end{figure}

\begin{figure}
    \centering
    \includegraphics[width=0.9\linewidth]{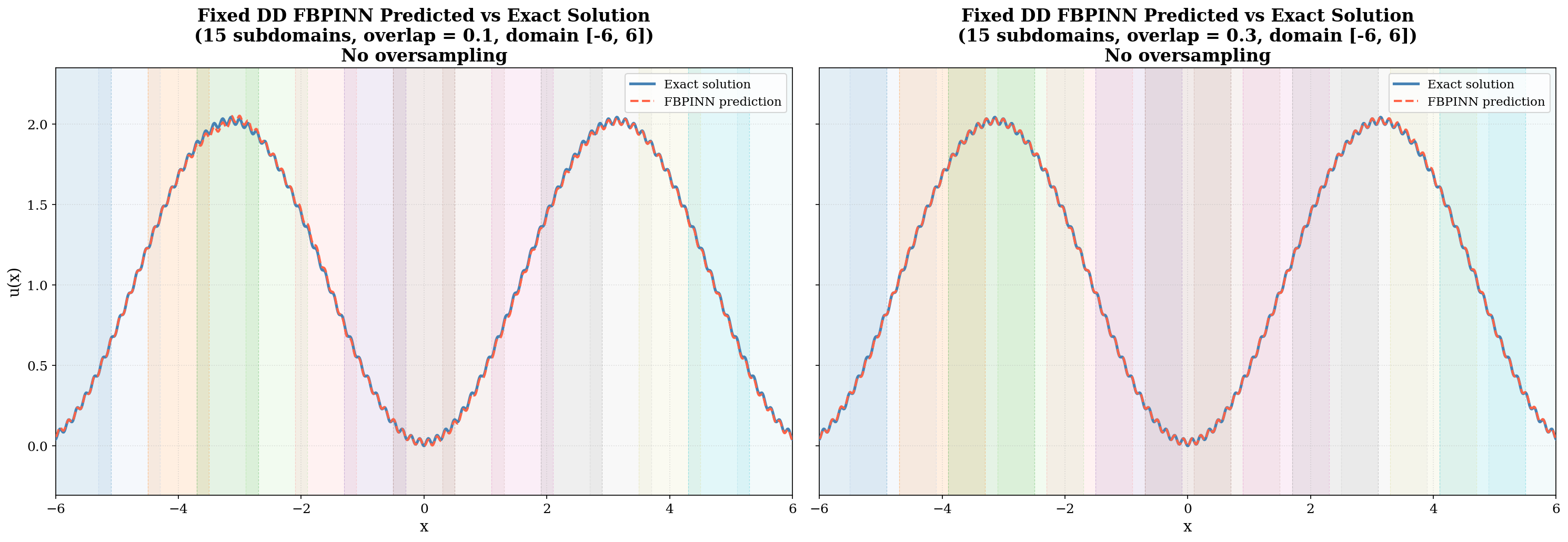}
    \caption{Predicted and exact results for the 1D Sine ODE with 2000 base points and an overlap width of $\delta=0.1$ (left) and $\delta=0.3$ (right). The wider overlap results in a more accurate solution.}
    \label{fig:compare_overlap}
\end{figure}

We first test this on the 1D Sine ODE in Equation \ref{eq:sineODE}. The total width of the overlap between two adjacent subdomains is set via a parameter $\delta,$ as described in Equation \ref{eq:overlap}. As the overlap width increases, the relative L2 error decreases, as seen in Table \ref{tab:sin_abl_errs}. On the other hand, simply adding collocation points to the overlapping regions does not significantly improve accuracy. In fact, we see that when the overlap is wider, $\delta=0.3,$ we still see a reduction in error from the thinner overlap $\delta=0.1$ even though the oversampled $\delta = 0.1$ case contains more overlap points. For our method, this means the adaptive domain decomposition must capture a wide enough overlap to ensure high accuracy with the FBPINN. In Fig. \ref{fig:abl_compare_oversampling}, we plot the predicted and exact solutions for the cases with and without oversampling in the overlap. As the errors suggest, oversampling does not correct the mismatch in dynamics. We also visualize the collocation points in the oversampled case in Fig. \ref{fig:sin1d_pts}. In contrast, Fig. \ref{fig:compare_overlap} depicts the predicted solutions with narrow overlap $\delta=0.1$ and wide overlap $\delta=0.3,$ and the improvement in the prediction from an FBPINN with wider overlapping subdomains is clear.

\begin{figure}[htbp]
    \centering
    \includegraphics[width=0.5\textwidth]{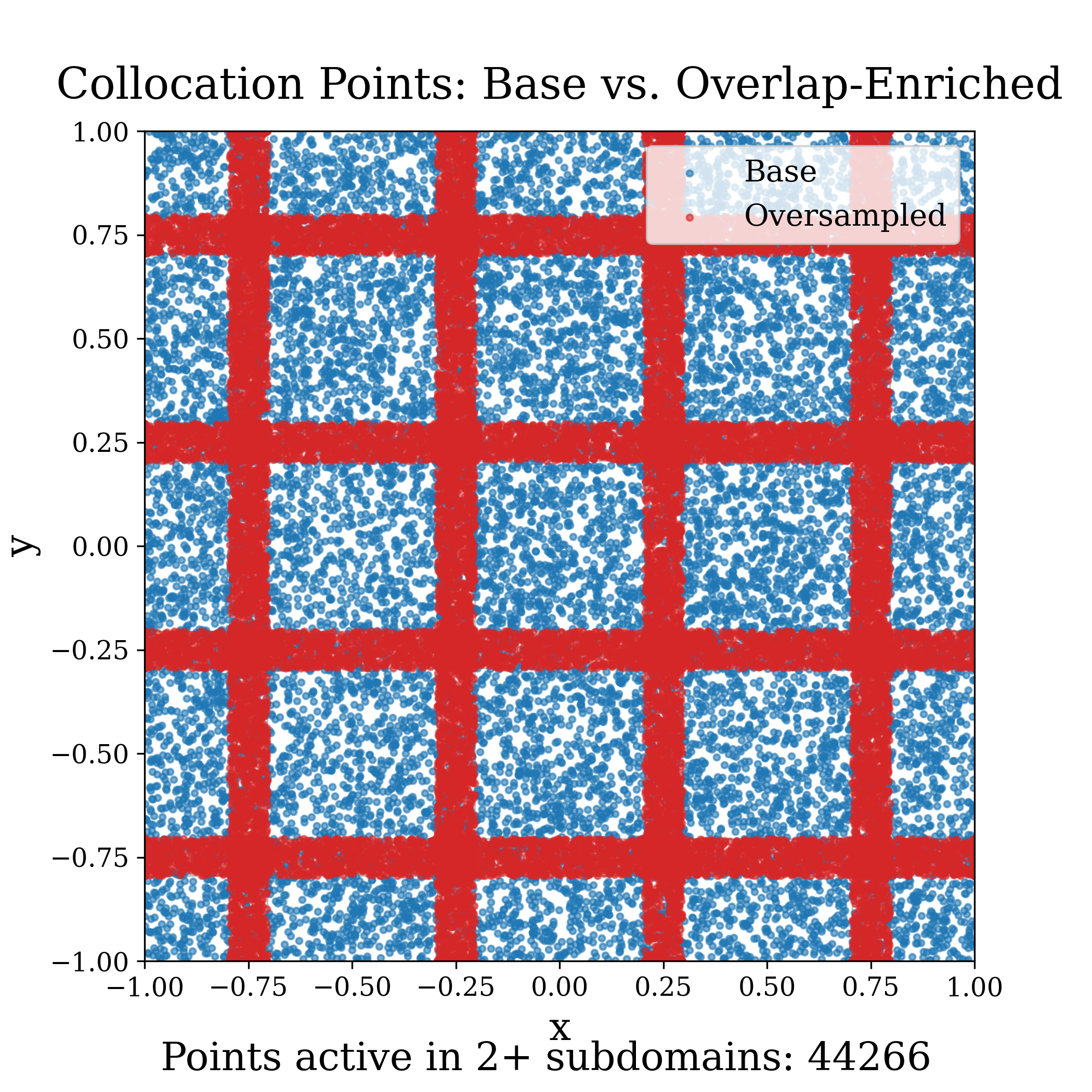}
    \caption{Distribution of collocation points for the 2D Helmholtz equation ablation study with an overlap width of $\delta=0.09$. 10000 base collocation points are sampled over the entire domain using Hammersley sampling. Then, extra points are added to regions where 2 or more subdomains are active with an oversampling factor of $\alpha=4.0.$}
    \label{fig:Helm_abl_points}
\end{figure}

Overlap width also affects accuracy for the 2D examples, with errors for the Helmholtz equation depicted in Table \ref{tab:Helmholtz_abl_errs} and the 2D Poisson equation with multiscale features in Table \ref{tab:Poisson2D_abl_errs}. For Helmholtz, when the points are oversampled in a narrow overlapping region no meaningful improvement in accuracy is seen. For this case, we also plot the distribution of collocation points for the cases with narrow overlap with and without oversampling in the overlap region in Fig. \ref{fig:Helm_abl_points}. Wider overlapping regions can reduce the error, whereas simply adding collocation points does not reliably improve accuracy. However, further increasing the overlap width need not improve accuracy, as the Poisson results show.

With these results, we conclude that overlap width rather than the density of collocation points in overlapping regions is a dominant factor controlling the accuracy with FBPINNs. This is in alignment with the findings in \cite{moseley2023fbpinns} and has a direct implication for our adaptive domain decomposition framework: it is insufficient for the POUNet to simply learn the number and placement of subdomains. The learned partitions must also exhibit wide enough overlapping regions to ensure smooth blending between subdomains, which results in higher accuracy for the downstream FBPINN training. However, the POUNet architectures naturally reward sharp partitions, since sharp boundaries reduce the approximation error in the POUNet's own training objective (Equation \ref{eq:POUobj}), because points active in multiple subdomains are represented by a blend of local polynomial models from the two subdomains which may disagree and increase error. Left unconstrained, the POUNet learns sharp partitions with insufficient overlap widths that are well suited for function approximation but inadequate for FBPINN training.  The adaptive domain decomposition results in the following section demonstrate that the Laplacian penalty regularization term increases the overlap regions between subdomains to allow for effective FBPINN training.

\begin{table}[htbp]

    \begin{center}
        \begin{tabular}{||c|c|c||}
            \hline
            {Overlap $\delta$} & Relative L2 error    & Points active in 2 subdomains \\ \hline
            $\delta=0.1$       & $7.38\times 10^{-3}$ & 533                                            \\
            \hline
            $\delta=0.1$       & $3.89\times 10^{-2}$ & 2206                                           \\
            \hline
            $\delta = 0.2$     & $4.67\times 10^{-3}$ & 1005                                           \\
            \hline
            $\delta = 0.3$     & $3.59\times10^{-3}$  & 1509                                           \\
            \hline
        \end{tabular}
    \end{center}
    \caption{Relative L2 errors and number of collocation points active in exactly 2 subdomains for the 1D Sine equation with $\omega_1=1$ and $\omega_2=45$ across a range of overlap widths $\delta = 0.1, 0.2, 0.3$ and $15$ uniform subdomains. The second row denotes a case where more collocation points were placed in the overlapping regions with an enrichment factor of 4.0, showing that more collocation points does not substantially boost accuracy.} \label{tab:sin_abl_errs}
\end{table}

\begin{table}[htbp]

    \begin{center}
        \begin{tabular}{||c|c|c||}
            \hline
            {Overlap $\delta$} & Relative L2 error    & Points active in 2 subdomains \\ \hline
            $\delta=0.09$      & $1.46\times 10^{-2}$ & 3353                                           \\
            \hline
            $\delta=0.09$      & $1.96\times 10^{-2}$ & 42,346                                         \\
            \hline
            $\delta = 0.14$    & $1.29\times 10^{-2}$ & 4887                                           \\
            \hline
            $\delta = 0.19$    & $5.52\times 10^{-3}$ & 6225                                           \\
            \hline
            $\delta = 0.24$    & $1.50\times10^{-3}$  & 7370                                           \\
            \hline
        \end{tabular}
    \end{center}
    \caption{Relative L2 errors and number of collocation points active in exactly 2 subdomains for the 2D Helmholtz equation with $a_1=1$ and $a_2=4$ across a range of overlap widths $\delta = 0.09, 0.14, 0.19, 0.24$ and $25$ uniform subdomains. The second row denotes a case where more collocation points were placed in the overlapping regions with an enrichment factor of 4.0, showing that more collocation points does not substantially boost accuracy.} \label{tab:Helmholtz_abl_errs}

\end{table}

\begin{table}[htbp]
    \centering
    \begin{tabular}{||c|c|c||}
        \hline
        {Overlap $\delta$} & Relative L2 error    & Points active in 2 subdomains \\
        \hline
        $\delta=0.09$      & $1.53\times10^{-1}$  & 7440                                           \\
        \hline
        \rowcolor{yellow!30}
        $\delta=0.09$      & $3.10\times10^{-1}$  & 44,266                                         \\
        \hline
        $\delta = 0.14$    & $8.33\times 10^{-2}$ & 8100                                           \\
        \hline
        $\delta = 0.19$    & $9.07\times 10^{-2}$ & 11,245                                         \\
        \hline
        $\delta = 0.24$    & $2.84\times 10^{-1}$ & 14,876                                         \\
        \hline
    \end{tabular}
    \caption{Relative L2 errors and number of collocation points active in exactly 2 subdomains for the 2D Poisson equation with multiscale features with $\omega_i=[1, 2, 4]$ across a range of overlap widths $\delta = 0.09, 0.14, 0.19, 0.24$ and $25$ uniform subdomains. An oversampling factor of $\alpha=4.0$ is used.}
    \label{tab:Poisson2D_abl_errs}
\end{table}

\subsection{ADD-FBPINN results}

In this section, we present the results for our ADD-FBPINN approach. All hyperparameters for each model can be found in Appendix \ref{sec:hyperparams}. The experiments are run on an NVIDIA A100 GPU. We use the relative $L_2$ error 

\begin{equation}\label{eq:L2}
    L_2 = \frac{\left(\sum_{j=1}^N\Vert\hat{u}(\mathbf{x}_j)-u(\mathbf{x}_j)\Vert^2\right)^{1/2}}{\left(\sum_{j=1}^N\Vert u(\mathbf{x}_j)\Vert^2\right)^{1/2}}
\end{equation}
\subsubsection{Helmholtz equation}

We first evaluate our ADD-FBPINN on the 2D Helmholtz equation presented in Equation \ref{eq:Helmholtzeq}. The Helmholtz equation is a classic benchmark for domain decomposition methods \cite{dolean2024mlfbpinn, jagtap2020xpinns} because its solution complexity is closely linked to the solution mode numbers $a_1$ and $a_2.$ Importantly, anisotropic cases where $a_1\neq a_2$ produce solutions with very different length scales along each axis, which uniform domain decomposition cannot efficiently capture with a small number of subdomains. And so, we use this example to show that our adaptive method can learn decompositions that are aligned with challenging regions of the solution.

We consider multiple cases by varying the mode number $a_1, a_2$ with examples ranging from low-to-high frequency, as well as isotropic and anisotropic cases. The cases are listed by mode numbers in Table \ref{tab:Helmholtz_errs} with relative L2 errors and number of subdomains learned for each case. For each case, we compare the relative L2 errors given by a baseline PINN, an FBPINN with fixed uniform domain decomposition, and the ADD-FBPINN. We select a range of overlap widths centered around the baseline overlap used in our main experiments. The fixed domain decomposition FBPINN always uses the same number of subdomains learned by the adaptive method to provide a fair comparison. For all cases, the separable RBF architecture is used for the POUNet, and the Laplacian penalty term with a weight of $\lambda_{\text{lap}}=0.0001$ is used, selected based on results from a preliminary sweep over $0.01, 0.001, 0.0001$ based on training stability and resulting overlap widths.

\begin{figure}[htbp]
    \centering
    \includegraphics[width=0.95\textwidth]{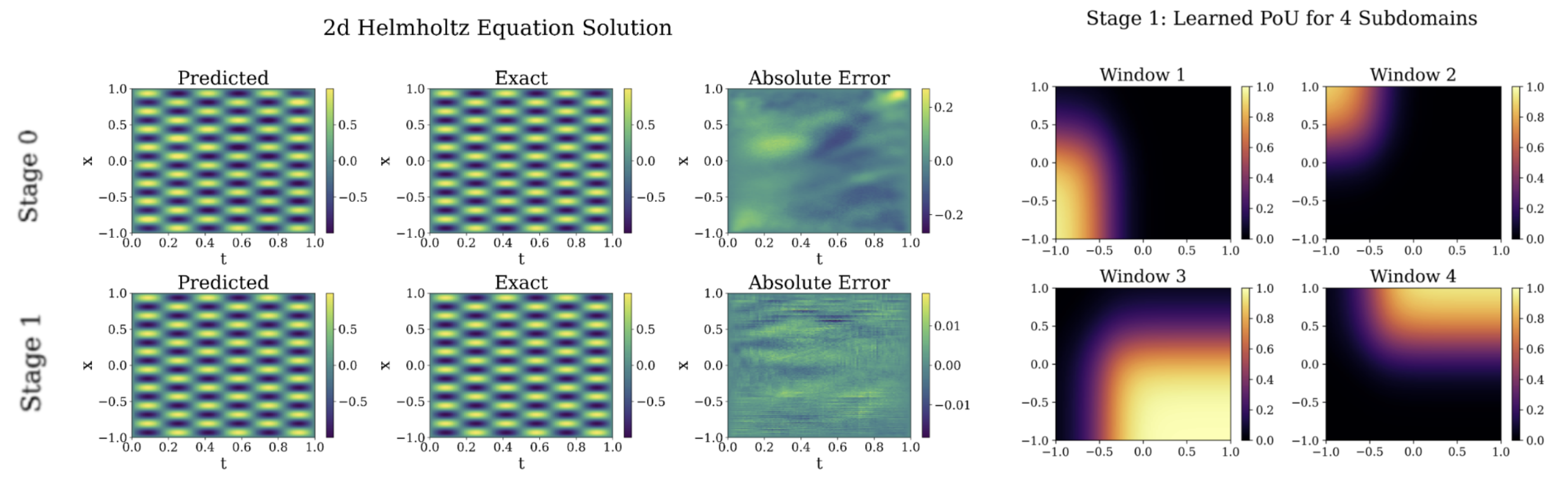}
    \caption{Results for the 2D Helmholtz equation with $a_1=3$ and $a_2=8.$ On the left, we see the predicted, exact, and absolute error for both Stage 0 (initial PINN), and Stage 1 (ADD-FBPINN). On the right, we see the learned subdomains used as the domain decomposition in Stage 1.}
    \label{fig:helm_3_8}

\end{figure}

The baseline PINN's performance drops in accuracy as the mode numbers increase, which is consistent with the well-documented failure mode of standard PINNs to capture the dynamics of high-frequency Helmholtz solutions. Our ADD-FBPINN reduces the error compared with fixed-decomposition FBPINNs in all cases, with improvements ranging from a few-fold to more than two orders of magnitude. Compared with standard PINNs, the error is reduced by more than one order of magnitude in all cases and by more than two orders of magnitude in Cases 3 and 4. In some cases, the learned domain decomposition is still mostly uniform, but shifted slightly to align with regions of high error. For instance, see the subdomains learned from Case 3 in Fig. \ref{fig:helm_3_8}. The partition locations correspond to regions of higher error given by the initial PINN prediction at Stage 0, by nature of the data-driven POUNet. The initial PINN has a lower error in the bottom right corner, so the POUNet places one large subdomain in that region, while the work in the higher-error regions is split by the remaining subdomains. This demonstrates that the adaptive framework does not simply impose a uniform domain decomposition strategy, but actually allocates subdomains in direct response to where the baseline solution is least accurate.

\begin{figure}[htbp]
    \centering
    \includegraphics[width=0.95\linewidth]{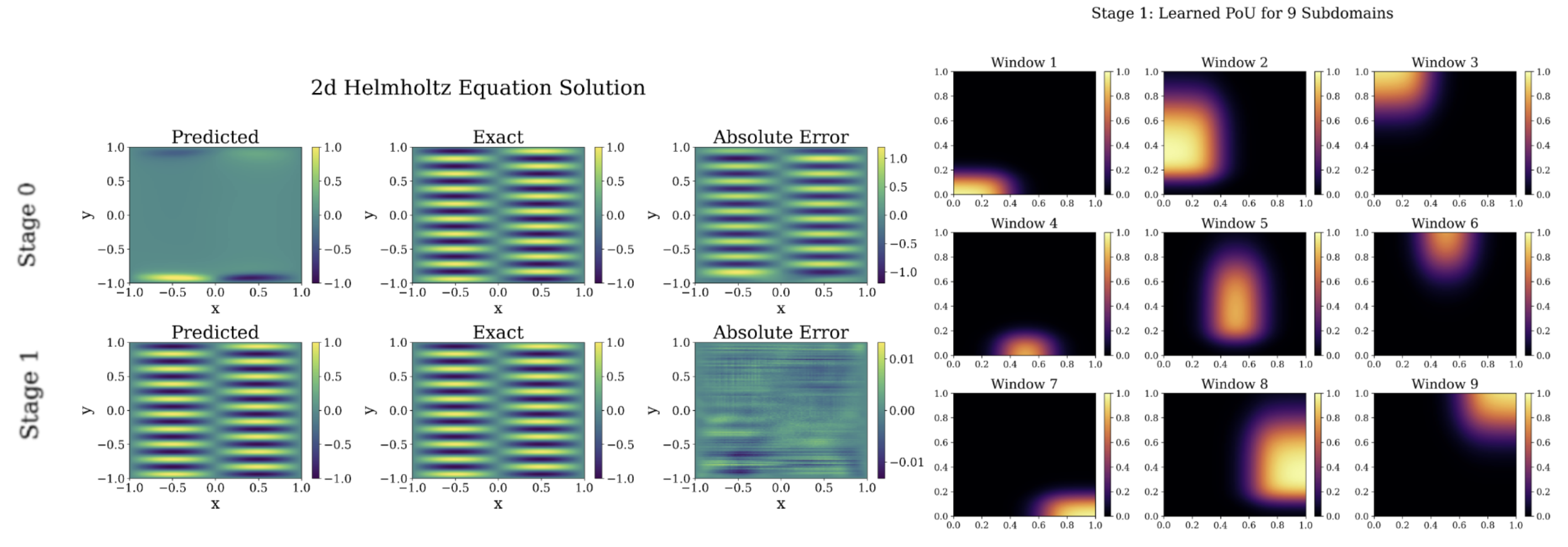}
    \caption{Results for the 2D Helmholtz equation with $a_1=1$ and $a_2=9.$ On the left, we see the predicted, exact, and absolute error for both Stage 0 (initial PINN), and Stage 1 (ADD-FBPINN). On the right, we see the learned subdomains used as the domain decomposition in Stage 1.}
    \label{fig:helm19}
\end{figure}

The largest relative gains relative to the baseline PINN are found in the anisotropic cases, where $a_1=3, a_2=8$ and $a_1=1, a_2=9.$ In particular, the baseline PINN completely fails to capture any meaningful dynamics in the highly anisotropic Case 4, but the adaptive framework is still able to recover and produce an accurate solution. The evolution of predictions from Stage 0, the initial PINN, to Stage 1, the FBPINN with learned subdomains, for Case 4 are depicted in Fig. \ref{fig:helm19}. Rather than uniform subdomains, the POUNet adapts the size and placement of each subdomain to the local complexity of the solution, as seen in Fig. \ref{fig:helm19}.

It is also worth noting that the fixed domain decomposition FBPINN, while outperforming the baseline PINN in most cases, does not achieve competitive accuracy in most cases. This is partly due to the fact that we constrain the fixed domain decomposition FBPINN to the same number of subdivisions learned by the adaptive method, which is often a small number. Beyond subdomain count, the fixed domain decomposition is not guaranteed to align with the length scales of the solution.  If more subdomains are used, the fixed decomposition errors improve at the cost of significant increase in computational cost. This highlights a key advantage of our approach: by learning subdomain placements aligned with features of the solution, the adaptive method achieves high accuracy with fewer subdomains. The POUNet only requires 2000 iterations to train, which is negligible compared to the cost of training an FBPINN on a larger number of domains.

\begin{table}[htbp]

    \begin{center}
        \begin{tabular}{||c|c|c|c|c|c||}
            \hline
            Case   & Mode numbers   & PINN                                 & FBPINN              & ADD-FBPINN           & Number of subdomains \\ \hline
            Case 1 & $a_1=1, a_2=4$ & $2.35\times 10^{-2}$                 & $1.50\times10^{-3}$ & $5.88\times10^{-4} $ & $4$                  \\
            \hline
            Case 2 & $a_1=6, a_2=6$ & $7.89\times10^{-2}$                  & $2.54\times10^{-1}$ & $6.41\times 10^{-3}$ & $4$                  \\
            \hline
            Case 3 & $a_1=3, a_2=8$ & $9.89\times10^{-1}$                  & $4.55\times10^{-1}$ & $3.89\times 10^{-3}$ & $4$                  \\
            \hline
            Case 4 & $a_1=1, a_2=9$ & $9.90\times10^{-1}$ & $6.02\times10^{-2}$ & $4.73\times10^{-3}$  & $9$                  \\
            \hline\end{tabular}

    \end{center}
    \caption{Relative L2 errors for a standard PINN, FBPINN, and ADD-FBPINN on the 2D Helmholtz equation. Results are shown for a range of solution mode numbers to highlight the ability of this method to capture both high-frequency isotropic cases and high-frequency anisotropic cases.} \label{tab:Helmholtz_errs}
\end{table}

\subsubsection{Poisson equation in two dimensions with multiscale features}

We use the 2D Poisson equation from Equation \ref{eq:Poissoneq} to demonstrate the ability of adaptive domain decomposition to capture multiscale features. Unlike the Helmholtz equation, where frequency is determined by a pair of solution mode numbers, the Poisson equation is constructed as a superposition of $n$ modes of frequencies $\{\omega_i\}_{i=1}^n,$ like in \cite{dolean2024mlfbpinn}. This allows us to directly control the spread and number of scales present in the solution. We consider four cases of increasing frequency, denoted in \ref{tab:Poisson_errs}, and test our ADD-FBPINN against a baseline PINN and fixed DD FBPINN.

\begin{figure}[htbp]
    \centering
    \includegraphics[width=0.95\linewidth]{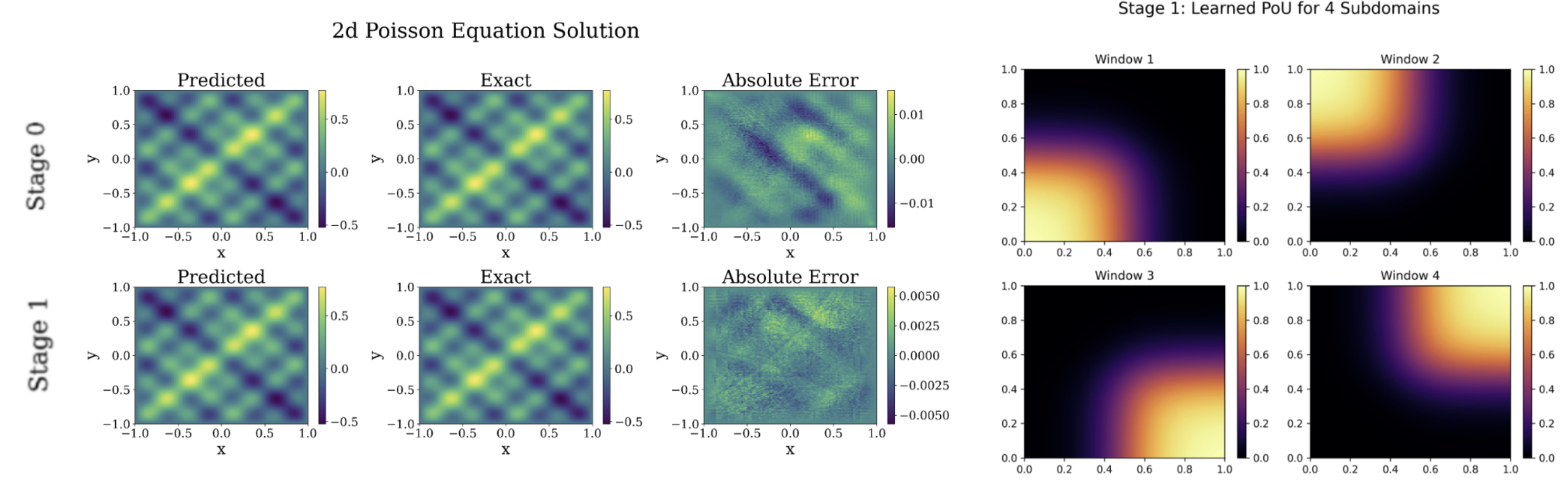}
    \caption{Results for the 2D Poisson equation with multiscale features for Case 3 with $\omega=[1,2,4,8].$ On the left, we have predicted, exact, and absolute error for both Stage 0 (baseline PINN) and Stage 1 (ADD-FBPINN). On the right, we see the learned domain decomposition used in Stage 1. In this case, the subdivision is still uniform, but the adaptive method learns the overlap width.}
    \label{fig:poisson_8}
\end{figure}

Final relative L2 errors are presented in Table \ref{tab:Poisson_errs}. In Cases 1 and 2, both the fixed and ADD-FBPINNs offer modest improvements over the baseline PINN, which is consistent with the expectation that domain decomposition provides the most benefit as solution complexity increases. In Case 3, the fixed DD FBPINN only slightly outperforms the baseline PINN despite using the same number of subdomains as the adaptive method. For Cases 1-3, the adaptive approach learns 4 subdomains that are more or less uniformly placed in the problem domain, similar to the partitions of the fixed DD FBPINN. So for these cases, placement is not the true gain of the adaptive approach, but learned overlap width. While overlap width can be selected according to the problem in a standard FBPINN, the adaptive approach learns the overlap without directly tuning its width, using a selected Laplacian penalty weight, as seen in Fig. \ref{fig:poisson_8}.

\begin{figure}[htbp]
    \centering
    \includegraphics[width=0.95\linewidth]{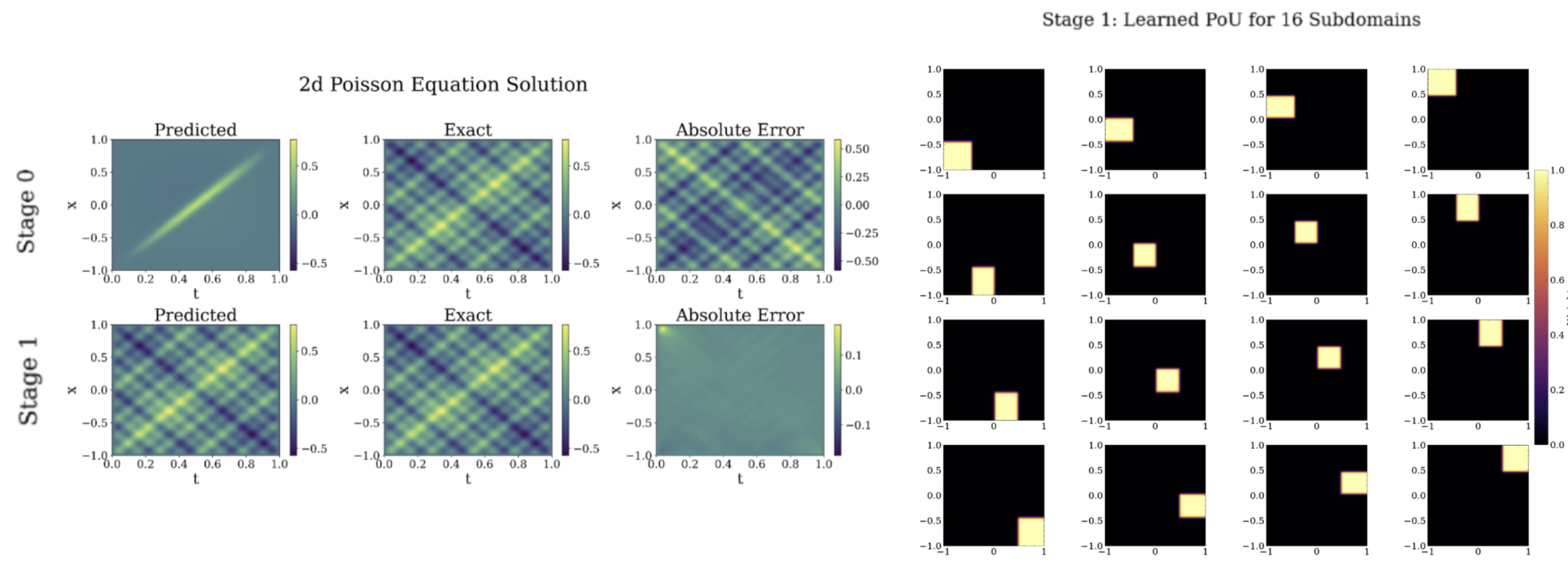}
    \caption{Results for the 2D Poisson equation with multiscale features for Case 4 with $\omega=[1,2,4,8, 16].$ On the left, we have predicted, exact, and absolute error for both Stage 0 (baseline PINN) and Stage 1 (ADD-FBPINN). On the right, we see the learned domain decomposition used in Stage 1. In this case, the  subdivision favors more subdomains with sharper transitions.}
    \label{fig:Poisson_16}

\end{figure}

A similar effect is present in Case 4 (Fig. \ref{fig:Poisson_16}), where the frequency is highest with $\omega_i=[1,2,4,8,16],$ the fixed FBPINN only slightly outperforms the baseline PINN, although both methods have higher errors than the ADD-FBPINNs. In this case, however, the ADD-FBPINN learns 16 subdivisions and favors a higher number of sharp partitions. 
The adaptive approach reduces the error by roughly an order of magnitude using the same subdomain count, by concentrating subdomains where the multiscale structure of the solution demands it.

\begin{table}[htbp]

    \begin{center}
        \begin{tabular}{||c|c|c|c|c|c||}
            \hline
            Case   & Frequencies $\{\omega_i\}$ & PINN                 & FBPINN               & ADD-FBPINN           & Number of subdomains \\ \hline
            Case 1 & $\omega=[1,2]$             & $6.09\times10^{-4}$  & $2.89\times10^{-4}$  & $9.77\times10^{-5}$  & $4$                  \\
            \hline
            Case 2 & $\omega=[1,2,4]$           & $1.76\times10^{-3}$  & $1.12\times 10^{-3}$ & $7.82\times10^{-4}$  & $4$                  \\
            \hline
            Case 3 & $\omega=[1,2,4,8]$         & $1.45\times 10^{-2}$ & $1.33\times10^{-2}$  & $4.33\times 10^{-3}$ & $4$                  \\
            \hline
            Case 4 & $\omega=[1,2,4,8,16]$      & $8.42\times10^{-1}$  & $5.84\times10^{-1}$  & $4.51\times10^{-2}$  & $16$                 \\
            \hline
        \end{tabular}
    \end{center}
    \caption{Relative L2 errors for a standard PINN, FBPINN, and ADD-FBPINN on the 2D Poisson equation with multiscale features. Results are shown for a range of frequencies with training difficulty increasing for each case.} \label{tab:Poisson_errs}
\end{table}

\subsubsection{Burgers equation}

We use viscous Burgers equation to highlight the ability of the ADD-FBPINN to accurately subdivide a problem domain not only in space, but also in time. This example also demonstrates that the POUNet accurately captures challenging regions of the solution, specifically, the sharp gradient region that forms around $t\approx0.3.$ We utilize two test cases here: first, the standard benchmark version depicted in Equation \ref{eq:Burgers} with the sharp gradient located at $x=0.0.$ Additionally, we consider Equation \ref{eq:b_periodic}, a non-symmetric variant where the shock is located at $x=-0.25.$

\begin{figure}[htbp]
    \centering
    \begin{subfigure}{0.45\textwidth}
        \centering\includegraphics[width=\textwidth]{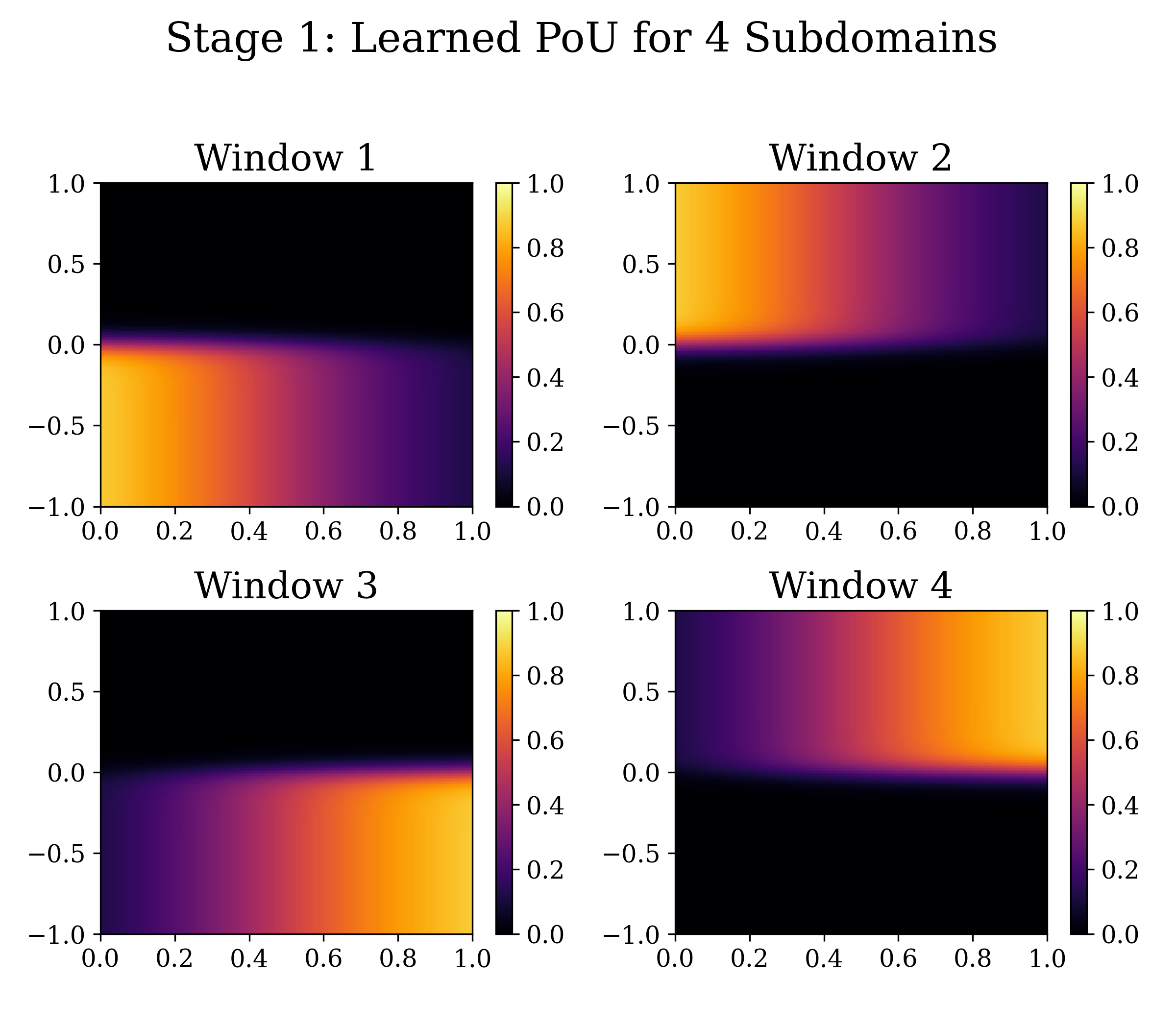}
        \caption{Subdomains learned for Burgers Equation \ref{eq:Burgers} with sharp gradient at $x=0.$}
    \end{subfigure}
    \hfill
    \begin{subfigure}{0.45\textwidth}
        \centering
        \includegraphics[width=\textwidth]{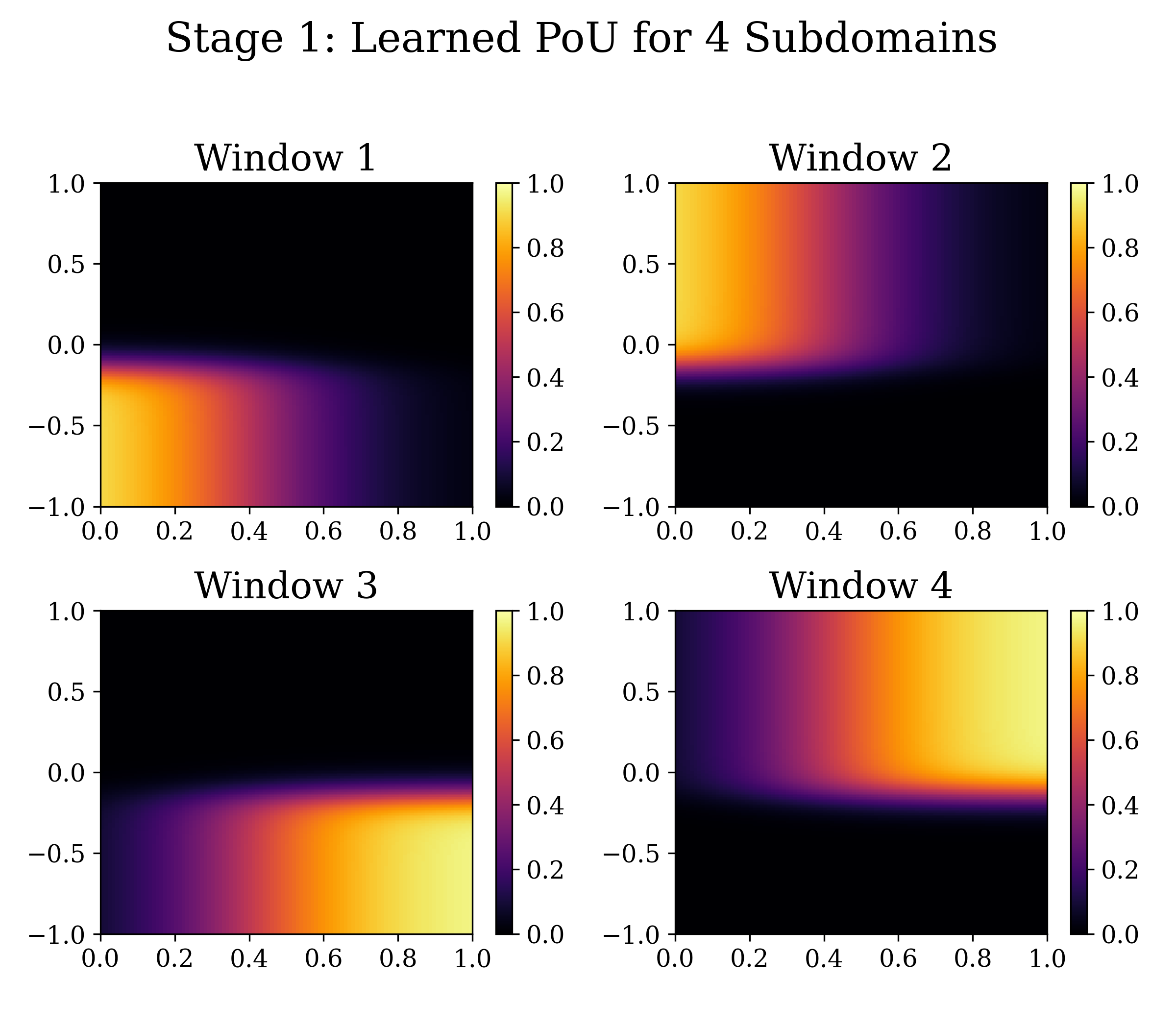}
        \caption{Subdomains learned for Burgers Equation \ref{eq:b_periodic} with sharp gradient at $x=-0.25.$}
        \label{fig:burgers_learned_periodic}
    \end{subfigure}
    \caption{Subdomains learned for various cases of Burgers equation by a POUNet.}
    \label{fig:burgers_learned_subdomains}
\end{figure}

We test these two cases to confirm that the POUNet does in fact learn subdomains that are aligned with challenging regions of the problem. In this case, the subdomains can locate the sharp gradient. The subdomains learned for each case are plotted in Fig. \ref{fig:burgers_learned_subdomains}. In fact, the POUNet for the periodic equation locates the position of the sharp gradient at $x=-0.25$ but also adds a wider overlap at the time where the shock forms $t\approx 0.3.$

\begin{figure}[htbp]
    \centering
    \includegraphics[width=0.9\linewidth]{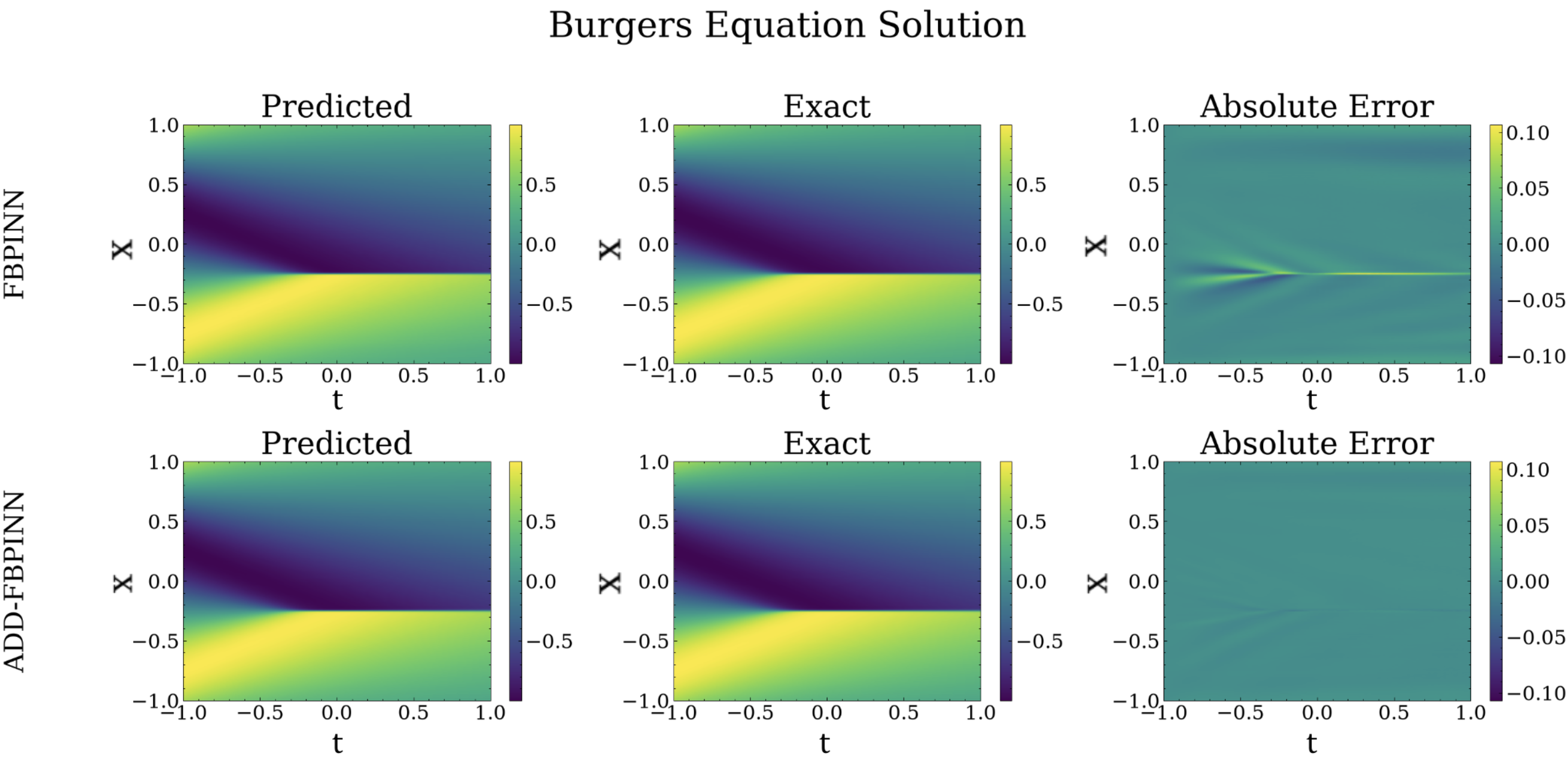}
    \caption{Predicted and exact solution and absolute error for Burgers equation comparing fixed domain decomposition with a standard FBPINN to that of an ADD-FBPINN.}
    \label{fig:Burgers_comparison_sol}

\end{figure}

\begin{figure}[htbp]
    \centering
    \begin{subfigure}{0.45\textwidth}
        \centering\includegraphics[width=\textwidth]{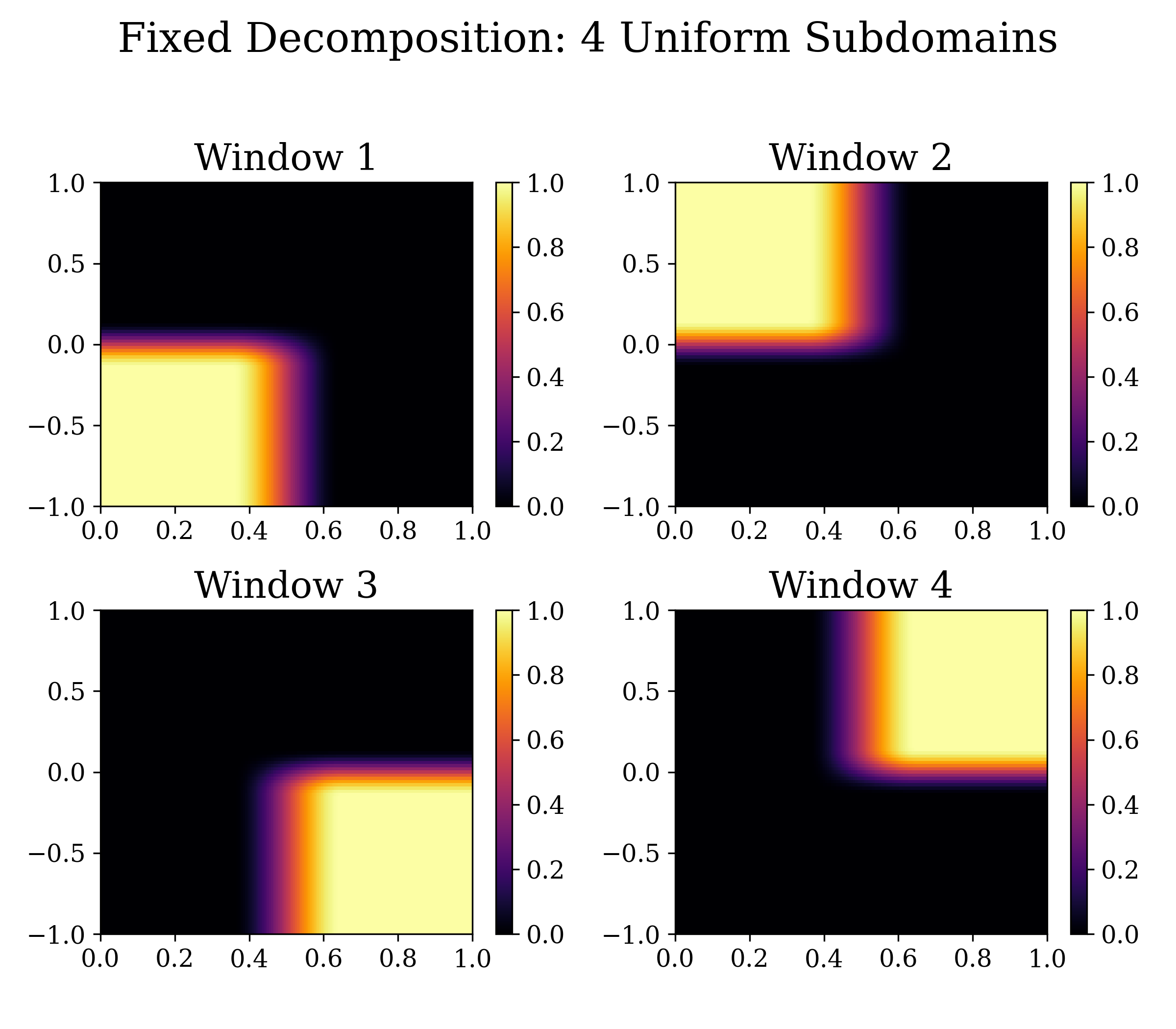}
        \caption{Fixed domain decomposition used for Burgers Equation \ref{eq:b_periodic} with sharp gradient at $x=-0.25.$}
    \end{subfigure}
    \hfill
    \begin{subfigure}{0.45\textwidth}
        \centering
        \includegraphics[width=\textwidth]{Figs/Burgers25.png}
        \caption{Subdomains learned for Burgers Equation \ref{eq:b_periodic} with sharp gradient at $x=-0.25.$}
        \label{fig:burgers_comparison_adaptive_periodic}

    \end{subfigure}
    \caption{Comparison between the fixed domain decomposition subdomains and the learned subdomains for Burgers equation with shock located at $x=-0.25.$ This case highlights the ability of the adaptive domain decomposition to perform feature-aligned domain decomposition automatically.}
    \label{fig:burgers_fixed_adaptive_subdomains}
\end{figure}

Relative L2 errors and number of subdomains learned are shown in Table \ref{tab:Burgers_errs}. Note that in the case where the shock is located at $x=-0.25,$ the baseline PINN even outperforms the FBPINN. We compare the predicted solutions and errors for this case in Fig. \ref{fig:Burgers_comparison_sol}. In Fig. \ref{fig:burgers_fixed_adaptive_subdomains}, we see that the subdomains the ADD-FBPINN learns are aligned with the challenging region of the problem, whereas the fixed DD subdivisions are uniform. The ADD-FBPINN can even find correct placement for subdomains  for problems where the difficulty arises in space-time. Our approach offers feature-aligned domain decomposition without the need to define such subdomains by hand for each problem of interest.

\begin{table}[htbp]

    \begin{center}
        \begin{tabular}{|c|c|c|c|c|}
            \hline
            Sharp gradient location & PINN                 & FBPINN               & ADD-FBPINN & Number of subdomains \\ \hline
            $x=0$                   & $9.85\times 10^{-2}$ & $4.59\times 10^{-2}$ & $1.26\times 10^{-3}$                & $4$                  \\
            \hline
            $x=-0.25$               & $1.48\times10^{-2}$  & $3.11\times 10^{-2}$ & $2.75\times10^{-3}$                 & $4$                  \\
            \hline
        \end{tabular}
    \end{center}
    \caption{Relative L2 errors for a standard PINN, FBPINN, and ADD-FBPINN on Burgers equation. Results are shown for two test cases, one where the shock is located at $x=0$ (Equation \ref{eq:Burgers}), and one where the shock is located at $x=-0.25$ (Equation \ref{eq:b_periodic})} \label{tab:Burgers_errs}
\end{table}

\section{Discussion}\label{sec:disc}

We have developed an adaptive domain decomposition framework for FBPINNs in which the window functions are learned rather than prescribed. POUNets are trained on an initial coarse PINN solution and the window functions learned are subsequently used as the subdomains for FBPINN training. No additional data are needed for the method, and it is purely physics-informed. We have introduced separable POUNets to enforce regularly shaped subdomains and a Laplacian penalty term to enforce wide enough overlapping regions between subdomains. Across three problems, the high-frequency 2D Helmholtz equation, multiscale 2D Poisson equation, and viscous Burgers equation, the adaptive method consistently outperforms a baseline PINN and fixed  FBPINN.

The ADD-FBPINN recognizes key features of the solution and proposes subdomains such that the challenging regions of the problem are shared across multiple subdomains. Further, with the Laplacian penalty term, the adaptive approach can efficiently find the overlap width between subdomains.

In future work we will consider adding local normalization of the input to the Laplacian penalty term, similar to the input normalization in FBPINNs that ensures the relative frequency each subdomain sees is low. We note that the size of the subdomains shrinks as the number of subdomains increases. When subdomains are smaller, the sharp transitions between subdomains learned by the POU loss occur over shorter distances, resulting in window functions having larger second derivatives. The Laplacian penalty term, then, grows as the size of subdomains decreases, which results in wider overlaps for small subdomains. At high subdomain counts, this can result in collapses of the subdomains. To further balance the sharpness enforced by the POUNet with the overlap regularization, one can normalize the input of the Laplacian penalty term to $[-1,1].$ Normalization allows the POUNet to learn a larger number of subdomains without overlaps swallowing the unique subdomain territory. This could be beneficial when the complexity of a solution demands a higher number of subdomains to reach sufficient accuracy, such as the highly anisotropic cases of the 2D Helmholtz problem.

Although the ADD-FBPINN framework can learn irregularly shaped domains with the MLP POUNet, regular subdomains yielded the highest FBPINN accuracy for the problems studied, likely due to the underlying PDEs having axis-aligned solution features. Additionally, the current implementation treats learned subdomains as a rectangle that covers the extent of the subdomain for normalization. This provides sufficient normalization for approximately rectangular subdomains, but may be less suitable for irregular subdomains. Developing a normalization approach that considers the geometry of an irregular subdomain is an additional direction for future work.

Another natural extension is to further automate the training hierarchy by replacing the user-defined schedule of partitions with an adaptive strategy that determines when and where additional subdivisions are needed during training.

\section{Acknowledgments}
The authors thank N. Trask for helpful discussions. 

The work of AH, SH, and PS is supported by the U.S. Department of Energy, Office of Science, Advanced Scientific Computing Research program under the Scalable, Efficient and Accelerated Causal Reasoning Operators, Graphs and Spikes for Earth and Embedded Systems (SEA-CROGS) project (Project No. 80278). The computational work was partially performed using PNNL Institutional Computing at Pacific Northwest National Laboratory. Pacific Northwest National Laboratory is a multi-program national laboratory operated for the U.S. Department of Energy by Battelle Memorial Institute under Contract No. DE-AC05-76RL01830.

During the preparation of this manuscript, the authors used ChatGPT Codex to improve language and readability. The authors reviewed and revised the resulting text and take full responsibility for the final manuscript.

\bibliography{bib.bib}
\bibliographystyle{plain}

\appendix
\section{MLP POUNets Architecture}\label{app:pou_arch}

In this section, we will provide details on the MLP POUNet architecture. An MLP is organized as a sequence of layers: an input layer, one or more hidden layers, and an output layer. 

The network comprises an input layer, $L$ hidden layers with widths $n_1, \ldots, n_L$ respectively, and an output layer. We consider an input vector $\*x\in\mathbb{R}^d$ and an output $\*y\in\mathbb{R}^{N_{\text{part}}}.$ These outputs parameterize the partition windows, whereas the PDE solution is scalar. Each hidden layer performs a weight sum plus bias, followed by a nonlinear activation function. Denoting the pre-activation and activation of the $k$-th hidden layer by $\*z^{(k)}$ and $\*a^{(k)},$ the neural network can be formulated as:
\begin{align*}
    \*z_j^{(1)}=\sum_{i=1}^{n_0}W_{ij}^{(1)}\*a_i^{(0)}+b_j^{(1)},       & \quad \*a_j^{(1)}=\sigma\left(\*z_j^{(1)}\right),               \\
    \*z_j^{(2)}=\sum_{i=1}^{n_1}W_{ij}^{(2)}\*a_i^{(1)}+b_j^{(2)},       & \quad \*a_j^{(2)}=\sigma\left(\*z_j^{(2)}\right),               \\
    \vdots                                                                                                                                 \\
    \*z_j^{(L)}=\sum_{i=1}^{n_{L-1}}W_{ij}^{(L)}\*a_i^{(L-1)}+b_j^{(L)}, & \quad \*a_j^{(L)}=\sigma\left(\*z_j^{(L)}\right),               \\
    y_j=\sum_{i=1}^{n_L}W_{ij}^{(L+1)}\*a_i^{(L)}+b_j^{(L+1)},           & \quad \*y=(y_1, \cdots y_{N_{\text{part}}})^T. \\
\end{align*}
The common choice for activation functions in MLPs is the Rectified Linear Unit (ReLU), defined as $\text{ReLU}(x)=\max(0,x),$ or hyperbolic tangent (tanh) function, given by $\text{tanh}(x)=\frac{e^x-e^{-x}}{e^x+e^{-x}}.$ In this work, we use the tanh activation function.

The neural network constitutes a composite function formed by these sequential layers. This function is parameterized by the weights $\*W^{(i)}$ and biases $\*b^{(i)},$ and can be expressed as a composition of functions:
\begin{align*}
    \mathcal{NN}(\*x;\theta) & =\*W^{(L+1)}\sigma(\*b^{(L)}+\*W^{(L)}\sigma(\cdots\sigma(\*b^{(1)}+\*W^{(1)}\*x)\cdots))+\*b^{(L+1)} \\
                             & =\*W^{(L+1)}\circ\*a^{(L)}\circ \cdots\circ \*a^{(1)}(\*x)+\*b^{(L+1)}.
\end{align*}
Here, $\theta=\{\*W^{(i)}, \*b^{(i)}\}_{i=1}^{L+1}$ is the set of all trainable parameters. We use the Glorot (Xavier) initialization \cite{glorot2010} to set the initial weights, with initial biases set to zero.

\section{Comparison of POUNet architectures}\label{sec:pou_comparison}

As presented in Section \ref{sec:methods}, multiple choices of POUNet architectures  can be used in this pipeline. In practice, the separable RBF POUNet learns subdivisions that result in the most accurate FBPINN predictions. As such, the results discussed in Section \ref{sec:results} all use the separable RBF POUNet. To provide motivations for this choice, in this section, we apply all four architectures to the benchmark 2D Helmholtz equation with $a_1=1, a_2=4.$

\begin{figure}[htbp]
    \centering
    \includegraphics[width=0.9\linewidth]{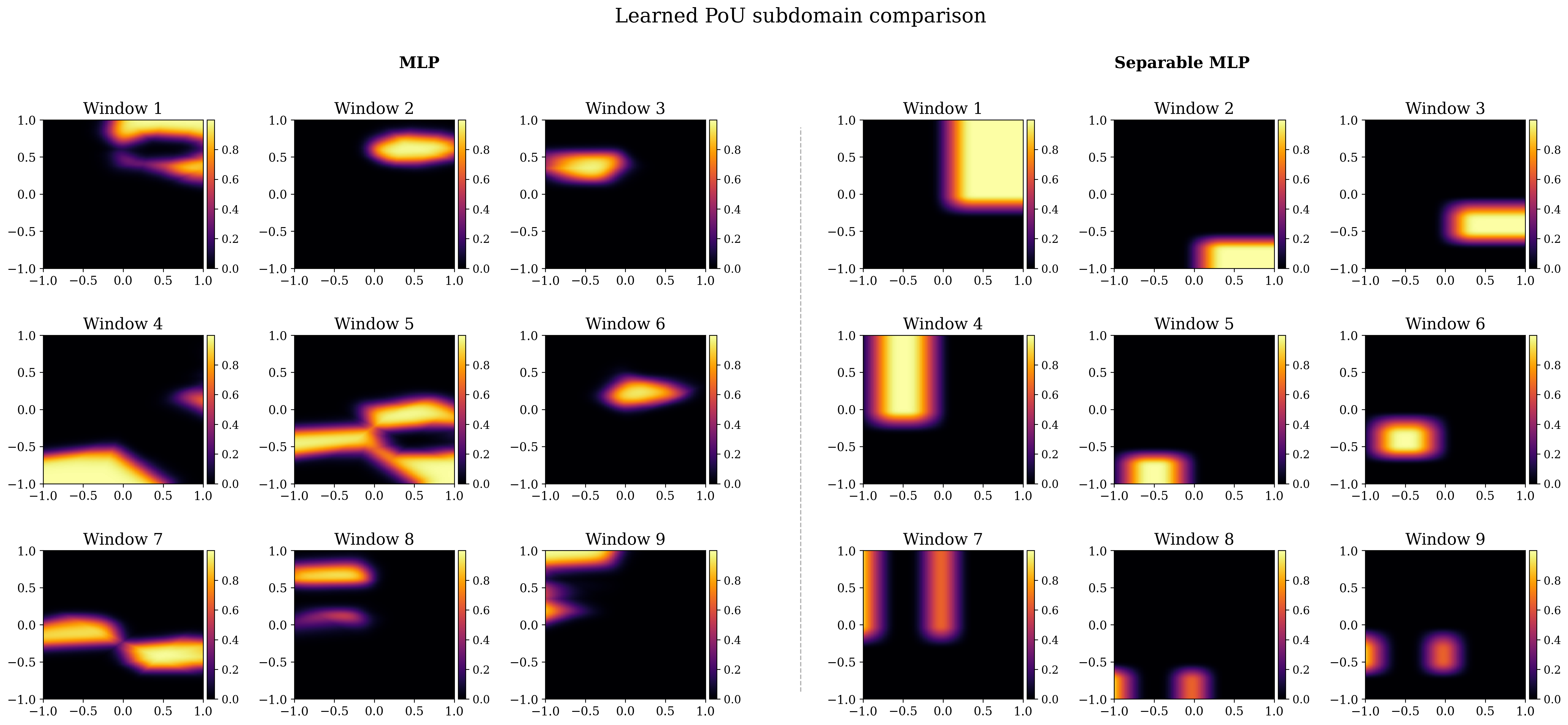}
    \caption{Learned domains given by MLP vs. Separable-MLP POUNet architectures.}
    \label{fig:mlp_comparison}
\end{figure}

\begin{figure}[htbp]
    \centering
    \includegraphics[width=0.9\linewidth]{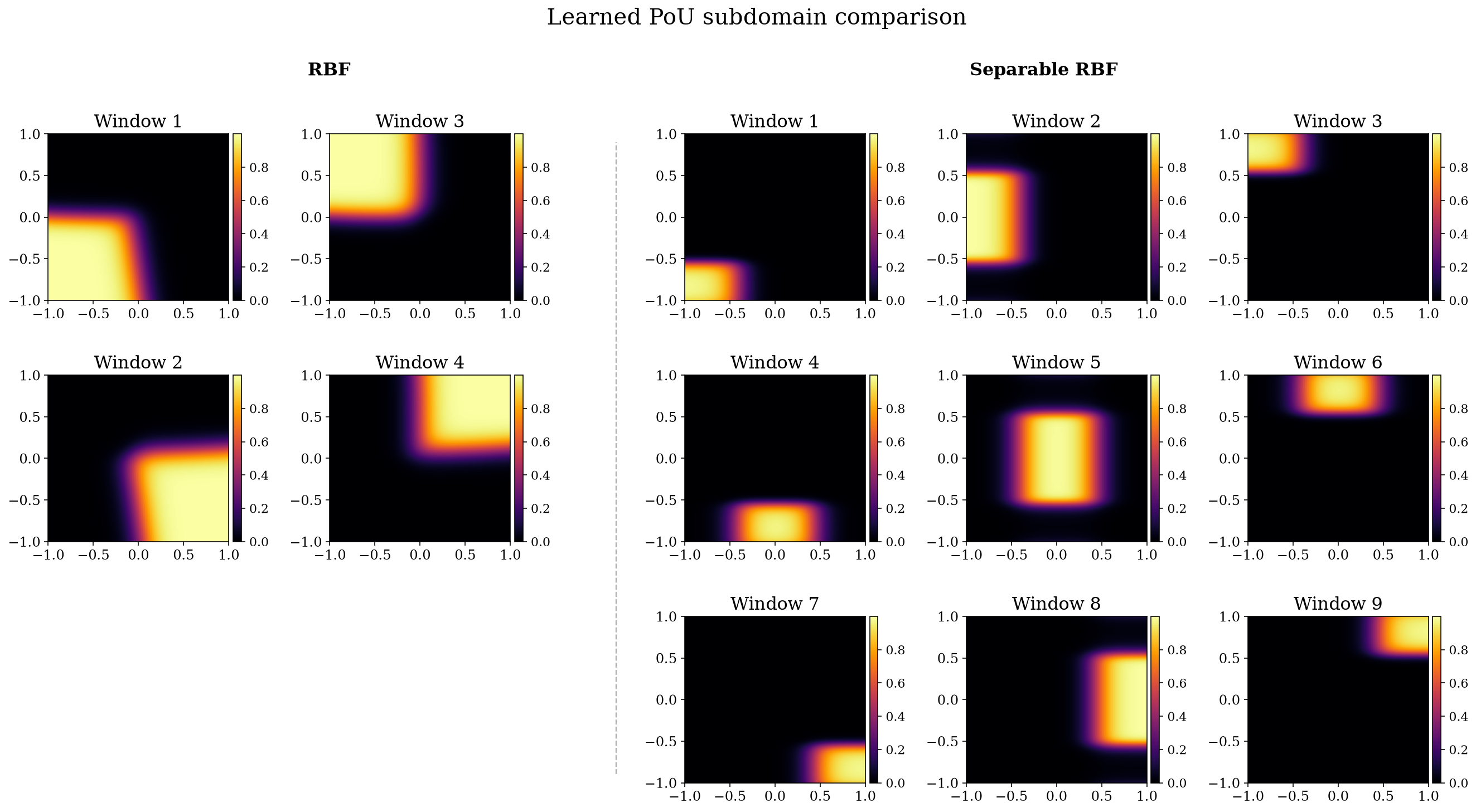}
    \caption{Learned domains given by RBF vs. Separable-RBF POUNet architectures.}
    \label{fig:rbf_comparison}
\end{figure}

In this test, we fix all model hyperparameters according to the best case results from Section \ref{sec:results}, and vary only the POUNet architecture. Hyperparameter settings can be found in Appendix \ref{sec:hyperparams}. The relative L2 errors for this test are depicted in Table \ref{tab:arch_comparison}. The subdomains learned with a Separable-RBF POUNet result in the lowest relative L2 error for downstream FBPINN training. Further, we plot comparison of all learned subdomains in Figs. \ref{fig:mlp_comparison} and \ref{fig:rbf_comparison}.

Note that, this test runs the full training pipeline, so number of subdomains learned can and does vary across architectures. Namely, the RBF POUNet has window functions that collapse at $9$ subdivisions, and so is unable to support as fine scale a partitioning as the other architectures.

\begin{table}[htbp]
    \centering
    \begin{tabular}{||c|c|c||}
        \hline
        Architecture                   & Relative L2 Error    & Number of subdomains \\
        \hline
        MLP                            & $6.99\times10^{-2}$  & 9                    \\
        \hline
        Separable MLP & $1.54\times 10^{-3}$ & 9                    \\
        \hline
        RBF                            & $1.31\times10^{-3}$  & 4                    \\
        \hline
        \rowcolor{yellow!30}
        Separable RBF & $7.83\times10^{-4}$  & 9                    \\
        \hline
    \end{tabular}
    \caption{Relative L2 errors and number of learned subdomains for the 2D Helmholtz equation given by the ADD-FBPINN. Errors are shown for a range of POUNet architectures, namely MLP, Separable-MLP, RBF, and Separable-RBF.}
    \label{tab:arch_comparison}

\end{table}

\clearpage 

\section{Model Hyperparameters}\label{sec:hyperparams}

\begin{table}[htbp]
    \centering
    \caption{Ablation Study Parameters}
    \label{tab:hyperparameters}
    \begin{tabular}{|c|c|c|c|}
        \hline
        FBPINN Parameter                          & 1D Sine ODE & 2D Helmholtz & 2D Poisson \\ \hline
        Learning rate                             & $0.001$                      & $0.0001$                      & $0.0001$                    \\ \hline
        Network size                              & $[1, 64, 64, 64, 64, 1]$     & $[2, 64, 64, 64, 64, 1]$      & $[2, 64, 64, 64, 64, 1]$    \\ \hline
        Activation function                       & $\text{tanh}$                & $\text{tanh}$                 & $\text{tanh}$               \\ \hline
        Collocation points (baseline)                       & 2,000                        & 10,000                        & 10,000                      \\ \hline
        Batch size                                & 1,000                        & 10,000                        & 10,000                      \\ \hline
        Iterations                                & 40,000                       & 40,000                        & 40,000                      \\ \hline
        RAD resample every                        & 10,000                       & 10,000                        & 10,000                      \\ \hline
        Subdomains per dimension & 15                           & 5                             & 5                           \\ \hline
    \end{tabular}
\end{table}

\begin{table}[htbp]
    \centering
    \caption{PINN Parameters}
    \begin{tabular}{|c|c|c|c|}
        \hline
        PINN Parameters     & 2D Helmholtz & 2D Poisson & Burgers                  \\ \hline
        Learning rate       & $0.0001$                      & $0.0001$                    & $0.0001$                 \\ \hline
        Network size        & $[2, 64, 64, 64, 64, 1]$      & $[2, 64, 64, 64, 64, 1]$    & $[2, 64, 64, 64, 64, 1]$ \\ \hline
        Activation function & $\text{tanh}$                 & $\text{tanh}$               & $\text{tanh}$            \\ \hline
        Collocation points  & 40,000                        & 40,000                      & 40,000                   \\ \hline
        Batch size          & 10,000                        & 10,000                      & 10,000                   \\ \hline
        Iterations          & 40,000                        & 40,000                      & 40,000                   \\\hline
    \end{tabular}
    \label{tab:pinn_parameters}

\end{table}

\begin{table}[htbp]
    \centering
    \caption{Fixed DD FBPINN Parameters}
    \begin{tabular}{|c|c|c|c|}
        \hline
        FBPINN Parameters    & 2D Helmholtz & 2D Poisson & Burgers                  \\ \hline
        Learning rate        & $0.0001$                      & $0.0001$                    & $0.0001$                 \\ \hline
        Network size         & $[2, 64, 64, 64, 64, 1]$      & $[2, 64, 64, 64, 64, 1]$    & $[2, 64, 64, 64, 64, 1]$ \\ \hline
        Activation function  & $\text{tanh}$                 & $\text{tanh}$               & $\text{tanh}$            \\ \hline
        Collocation points   & 40,000                        & 40,000                      & 40,000                   \\ \hline
        Batch size           & 10,000                        & 10,000                      & 10,000                   \\ \hline
        Iterations           & 40,000                        & 40,000                      & 40,000                   \\\hline
        Number of subdomains & Varies                        & Varies                      & Varies                   \\\hline
    \end{tabular}
    \label{tab:fixed_fbpinn_parameters}
\end{table}
\begin{table}[htbp]
    \centering
    \caption{ADD-FBPINN Parameters}
    \begin{tabular}{|c|c|c|c|}
        \hline
        ADD-FBPINN Parameters    & 2D Helmholtz & 2D Poisson & Burgers                  \\ \hline
        Learning rate        & $0.0001$                      & $0.0001$                    & $0.0001$                 \\ \hline
        Network size         & $[2, 64, 64, 64, 64, 1]$      & $[2, 64, 64, 64, 64, 1]$    & $[2, 64, 64, 64, 64, 1]$ \\ \hline
        Activation function  & $\text{tanh}$                 & $\text{tanh}$               & $\text{tanh}$            \\ \hline
        Collocation points   & 40,000                        & 40,000                      & 40,000                   \\ \hline
        Batch size           & 10,000                        & 10,000                      & 10,000                   \\ \hline
        Iterations           & 40,000                        & 40,000                      & 40,000                   \\\hline
        RAD resampling every & 10,000                        & 10,000                      & 10,000                   \\ \hline
    \end{tabular}
\end{table}

\begin{table}[htbp]
    \centering
    \caption{ADD-FBPINN POUNet Parameters}
    \label{tab:pounet_parameters}
 
    \begin{tabular}{|c|c|c|c|}
        \hline
        ADD-FBPINN POUNet Parameters & 2D Helmholtz              & 2D Poisson                & Burgers                                    \\ \hline
        Polynomial degree & 2 & 2 & 2 \\
        \hline
        $N_{\text{data}}$ & 40,000 & 40,000 & 40,000 \\
        \hline
        
        Learning rate                      & $0.001$                                    & $0.001$                                    & $0.001$                                    \\ \hline
        Architecture                       & Separable RBF                              & Separable RBF                              & Separable RBF                              \\ \hline
        Iterations                         & 2,000                                      & 2,000                                      & 2,000                                      \\ \hline
        Collapse threshold                 & $\tau=0.8$                                 & $\tau=0.8$                                 & $\tau=0.8$                                 \\ \hline
        POU schedule                       & \makecell{[1,4,9,16,25,36,\\49,64,81,100]} & \makecell{[1,4,9,16,25,36,\\49,64,81,100]} & \makecell{[1,4,9,16,25,36,\\49,64,81,100]} \\ \hline
    \end{tabular}

\end{table}

\clearpage

\section{RAD resampling}\label{sec:rad}

To improve sampling of collocation points across the computational domain, we employ the residual-based adaptive distribution (RAD) resampling method \cite{wu2023rad}. The frequency of RAD resampling used in this work is noted in Appendix \ref{sec:hyperparams}.

At each resampling step, a new set of collocation points is drawn from a probability density function (PDF) $p(\*x)$ that is proportional to the current residual error. For any point $\*x\in\Omega,$ the residual is $\varepsilon(\*x)=\lvert\mathcal{D}[\hat{u}(\*x;\theta)]-f(\*x)\rvert.$ The sampling PDF is proportional to
\begin{equation}
    p(\*x)\propto\frac{\varepsilon(\*x)^k}{\mathbb{E}[\varepsilon(\*x)^k]}+c, \quad k\geq0, c\geq0,
\end{equation}
where expectation is taken over the uniform measure on $\Omega$ and is estimated numerically. The hyperparameters $k$ and $c$ are tuneable, and as $k$ increases, the sampling distribution becomes more concentrated in regions with larger residuals, allowing the optimizer to focus on areas that are more challenging to approximate. The parameter $c$ acts as a regularizer; as $c$ increases, the sampling distribution tends toward a uniform distribution. As suggested by \cite{wu2023rad}, we set $k=c=1$ as a default setting for our work.

\end{document}